\documentclass[11pt]{article}
\usepackage{acl2014}
\usepackage{times}
\usepackage{url}
\usepackage{graphicx,epstopdf}
\usepackage{latexsym}
\usepackage[tight]{subfigure}
\usepackage{algorithm}
\usepackage{algorithmic}
\usepackage{amsmath} 
\usepackage{amssymb}  
\usepackage{cases}
\usepackage{collref}
\usepackage{float}
\usepackage{array}
\usepackage{epsfig}
\usepackage{multicol}
\newcolumntype{C}[1]{>{\centering}p{#1}}

\DeclareMathOperator*{\argmin}{arg~min}

\title{Errata: Distant Supervision for Relation Extraction with Matrix Completion}

\author{Miao Fan$^{\dagger,\ddagger,*}$, Deli Zhao$^{\ddagger}$, Qiang Zhou$^{\dagger}$, Zhiyuan Liu$^{\diamond, \ddagger}$, Thomas Fang Zheng$^{\dagger}$, Edward Y. Chang$^{\ddagger}$\\
$^{\dagger}$ CSLT, Division of Technical Innovation and Development,\\ Tsinghua National Laboratory for Information Science and Technology, Tsinghua University, China.\\
$^{\diamond}$ Department of Computer Science and Technology, Tsinghua University, China. \\
$^{\ddagger}$ HTC Beijing Advanced Technology and Research Center, China.\\
{\tt $^*$fanmiao.cslt.thu@gmail.com}}

\date{}

\begin{document}

\twocolumn[
\begin{center} \LARGE{\textbf{Errata to our ACL'14 paper}} \end{center} \vspace{0.5cm}

{This errata is to correct the experiments of method comparison (Figure 4 and Figure 6) on NYT'10 dataset in our ACL'14 paper:
\vspace{0.5cm}

{\small{Miao Fan, Deli Zhao, Qiang Zhou, Zhiyuan Liu, Thomas Fang Zheng, Edward Y. Chang. 2014. Distant Supervision for Relation Extraction with Matrix Completion. In \emph{Proceedings of the 52nd Annual Meeting of the Association for Computational Linguistics} (ACL'14). }}\vspace{0.5cm}

We made a mistake that we tested the performance of the previous approaches (Mintz'09, MultiR'11, MIML'12 and MIML-at-least-one'12) with the dataset of NYT'10 which contains much more instances labeled by {\bf NG}, but plotted the precision-recall curve of our DRMC-b and DRMC-1 approaches without considering the {\bf NG}-labeled testing instances. \vspace{0.5cm}

To make up the erroneous experiments, we filter out the instances with {\bf NG} labels in the NYT'10 dataset and test the instances labeled by at least one positive relation in this paper. Table 1 shows the statistics of the training and testing sets. We re-run the experiments of method comparison among Mintz'09, MultiR'11, MIML'12, MIML-at-least-one'12, DRMC-b and DRMC-1, and show the results in the replaced figures (Figure 4 and Figure 6). Hopefully, our algorithms still significantly outperform the state-of-art methods. Therefore, the claims in our ACL'14 paper still hold. \vspace{0.5cm}

This errata paper can be cited as:

{\small{Miao Fan, Deli Zhao, Qiang Zhou, Zhiyuan Liu, Thomas Fang Zheng, Edward Y. Chang. 2014. Errata: Distant Supervision for Relation Extraction with Matrix Completion. In arXiv 1116094.}}\vspace{0.5cm}

For who are interested in our studies, please do refer to this updated version. Thanks so much to Danqi Chen in Stanford University who pointed out the mistakes. We sincerely apologize for the carelessness.\vspace{0.5cm}

In the future, we look forward to exploring better approaches on completing large-scale matrices that include more training instances with {\bf NG} labels for distantly supervised relation extraction task.\vspace{0.5cm}
}
]

\maketitle
\begin{abstract}
The essence of distantly supervised relation extraction is that it is an {\it incomplete} multi-label classification problem with {\it sparse} and {\it noisy} features. To tackle the sparsity and noise challenges, we propose solving the classification problem using matrix completion on factorized matrix of minimized rank.
We formulate relation classification as completing the unknown labels of testing items (entity pairs) in a sparse matrix that concatenates training and testing textual features with training labels.
Our algorithmic framework is based on the assumption that the rank of item-by-feature and item-by-label joint matrix is low. We apply two optimization models to recover the underlying low-rank matrix leveraging the sparsity of feature-label matrix. The matrix completion problem is then solved by the fixed point continuation (FPC) algorithm, which can find the global optimum. Experiments on two widely used datasets with different dimensions of textual features demonstrate that our low-rank matrix completion approach significantly outperforms the baseline and the state-of-the-art methods.
\end{abstract}

\section{Introduction}
Relation Extraction (RE) is the process of generating structured relation knowledge from unstructured natural language texts. Traditional supervised methods \cite{guodong2005exploring, Bach2007} on small hand-labeled corpora, such as MUC\footnote{http://www.itl.nist.gov/iaui/894.02/related projects/muc/} and ACE\footnote{http://www.itl.nist.gov/iad/mig/tests/ace/}, can achieve high precision and recall. However, as producing hand-labeled corpora is laborius and expensive, the supervised approach can not satisfy the increasing demand of building large-scale knowledge repositories with the explosion of Web texts.
To address the lacking training data issue, we consider the distant \cite{mintz2009distant} or weak \cite{hoffmann-EtAl:2011:ACL-HLT2011} supervision paradigm attractive, and we improve the effectiveness of the paradigm in this paper.

\begin{figure}
\centering
\includegraphics[width=0.47\textwidth]{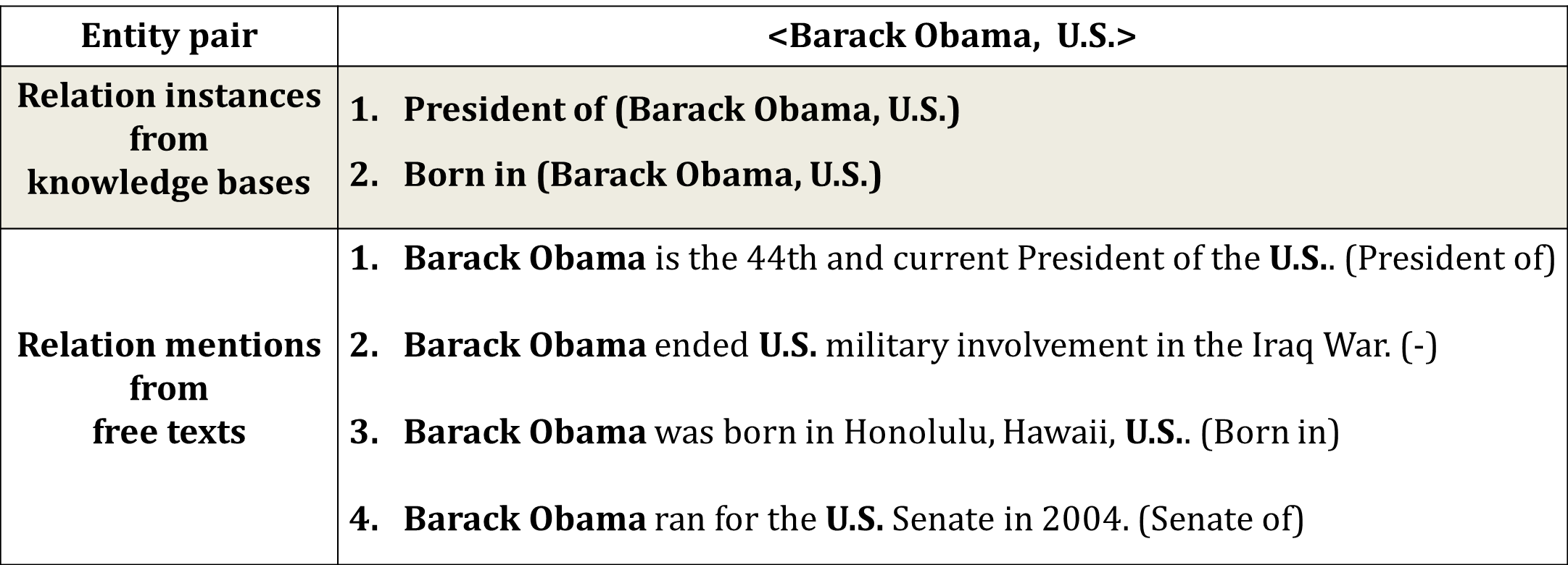}\\

\caption{Training corpus generated by the basic alignment assumption of distantly supervised relation extraction. The relation instances are the triples related to President Barack Obama in the Freebase, and the relation mentions are some sentences describing him in the Wikipedia.}

\end{figure}

The intuition of the paradigm is that one can take advantage of several knowledge bases, such as WordNet\footnote{http://wordnet.princeton.edu}, Freebase\footnote{http://www.freebase.com} and YAGO\footnote{http://www.mpi-inf.mpg.de/yago-naga/yago}, to automatically label free texts, like Wikipedia\footnote{http://www.wikipedia.org} and New York Times corpora\footnote{http://catalog.ldc.upenn.edu/LDC2008T19}, based on some heuristic alignment assumptions.
An example accounting for the basic but practical assumption is illustrated in Figure 1, in which we know that the two entities ($<${\tt Barack Obama, U.S.}$>$) are not only involved in the {\it relation instances\footnote{According to convention, we regard a structured triple $r(e_i, e_j)$ as a relation instance which is composed of a pair of entities \textless$e_i, e_j$\textgreater and a relation name $r$ with respect to them.}} coming from knowledge bases ({\tt President-of(Barack Obama, U.S.)} and {\tt Born-in(Barack Obama, U.S.)}), but also co-occur in several {\it relation mentions}\footnote{The sentences that contain the given entity pair are called relation mentions.} appearing in free texts ({\tt Barack Obama is the 44th and current President of the U.S.} and {\tt Barack Obama was born in Honolulu, Hawaii, U.S.}, etc.). We extract diverse textual features from all those {\it relation mentions} and combine them into a rich feature vector labeled by the {\it relation names} ({\tt President-of} and {\tt Born-in}) to produce a {\it weak} training corpus for relation classification.
\begin{figure*}[htb]
\centering
\includegraphics[width=1.05\textwidth]{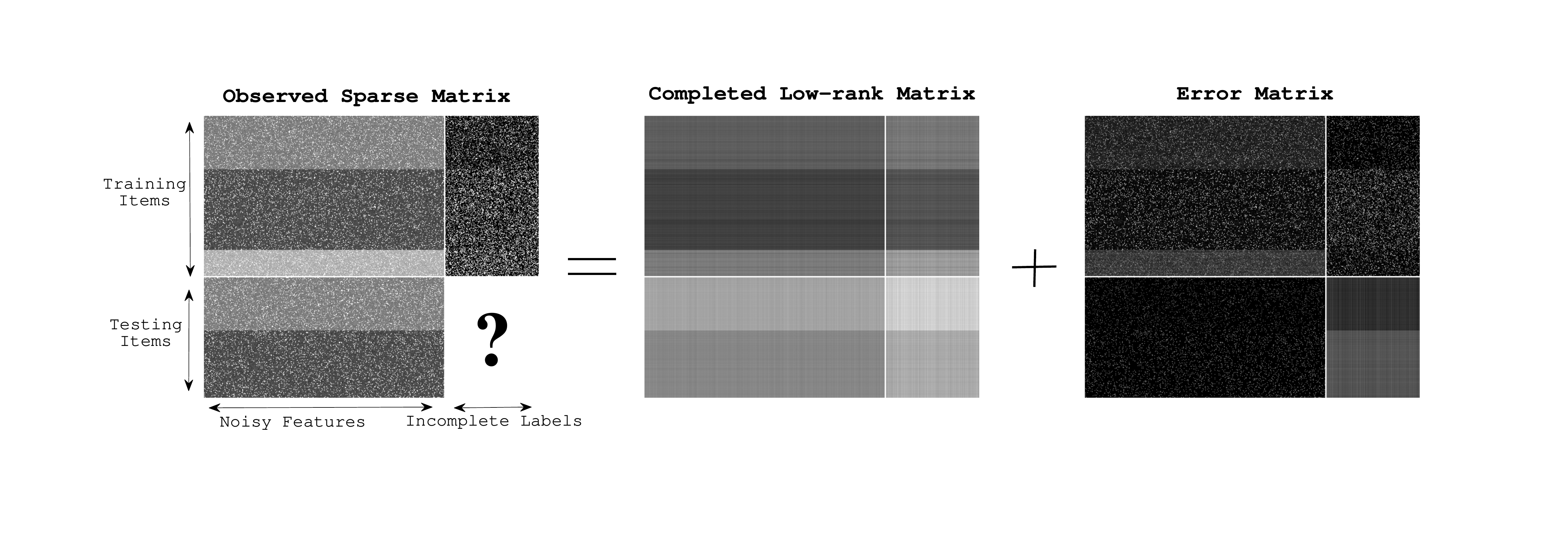}\\
\vspace{-30pt}
\caption{The procedure of noise-tolerant low-rank matrix completion. In this scenario, distantly supervised relation extraction task is transformed into completing the labels for testing items (entity pairs) in a sparse
matrix that concatenates training and testing textual features with training labels. We seek to recover the underlying low-rank matrix and to complete the unknown testing labels simultaneously.}
\vspace{10pt}
\end{figure*}

This paradigm is promising to generate large-scale training corpora automatically. However, it comes up against three technical challeges:
\begin{itemize}

\item {\bf Sparse features}. As we cannot tell what kinds of features are effective in advance, we have to use NLP toolkits, such as Stanford CoreNLP\footnote{http://nlp.stanford.edu/downloads/corenlp.shtml}, to extract a variety of textual features, e.g., named entity tags, part-of-speech tags and lexicalized dependency paths. Unfortunately, most of them appear only once in the training corpus, and hence leading to very sparse features.

\item {\bf Noisy features}. Not all relation mentions express the corresponding relation instances. For example, the second relation mention in Figure 1 does not explicitly describe any relation instance, so features extracted from this sentence can be noisy. Such analogous cases commonly exist in feature extraction.

\item {\bf Incomplete labels}. Similar to noisy features, the generated labels can be incomplete. For example, the fourth relation mention in Figure 1 should have been labeled by the relation {\tt Senate-of}. However, the incomplete knowledge base does not contain the corresponding relation instance ({\tt Senate-of(Barack Obama, U.S.)}). Therefore, the distant supervision paradigm may generate incomplete labeling corpora.
\end{itemize}
In essence, distantly supervised relation extraction is an {\it incomplete} multi-label classification task with {\it sparse} and {\it noisy} features.

In this paper, we formulate the relation-extraction task from a novel perspective of using matrix completion with low rank criterion. To the best of our knowledge, we are the first to apply this technique on relation extraction with distant supervision.
More specifically, as shown in Figure 2, we model the task with a sparse matrix whose rows present items (entity pairs) and columns contain noisy textual features and incomplete relation labels. In such a way, relation classification is transformed into a problem of completing the unknown labels for testing items in the sparse matrix that concatenates training and testing textual features with training labels, based on the assumption that the item-by-feature and item-by-label joint matrix is of low rank.
The rationale of this assumption is that noisy features and incomplete labels are semantically correlated. The low-rank factorization of the sparse feature-label matrix delivers the low-dimensional representation of de-correlation for features and labels.

We contribute two optimization models, DRMC\footnote{It is the abbreviation for {\bf D}istant supervision for {\bf R}elation extraction with {\bf M}atrix {\bf C}ompletion}-b and DRMC-1, aiming at exploiting the sparsity to recover the underlying low-rank matrix and to complete the unknown testing labels simultaneously. Moreover, the logistic cost function is integrated in our models to reduce the influence of noisy features and incomplete labels, due to that it is suitable for binary variables. We also modify the fixed point continuation (FPC) algorithm \cite{Ma2011} to find the global optimum.

Experiments on two widely used datasets demonstrate that our noise-tolerant approaches outperform the baseline and the state-of-the-art methods. Furthermore, we discuss the influence of feature sparsity, and our approaches consistently achieve better performance than compared methods under different sparsity degrees.
\section{Related Work}
The idea of distant supervision was firstly proposed in the field of bioinformatics \cite{Craven1999}. Snow et al. \shortcite{Snow2004} used WordNet as the knowledge base to discover more hpyernym/hyponym relations between entities from news articles. However, either bioinformatic database or WordNet is maintained by a few experts, thus hardly kept up-to-date.

As we are stepping into the {\em big data} era, the explosion of unstructured Web texts simulates us to build more powerful models that can automatically extract relation instances from large-scale online natural language corpora without hand-labeled annotation.
Mintz et al. \shortcite{mintz2009distant} adopted Freebase \cite{Bollacker2008, Bollacker2007}, a large-scale crowdsourcing knowledge base online which contains billions of relation instances and thousands of relation names, to {\it distantly supervise} Wikipedia corpus. The basic alignment assumption of this work is that if a pair of entities participate in a relation, {\it all sentences} that mention these entities are labeled by that relation name. Then we can extract a variety of textual features and learn a multi-class logistic regression classifier. Inspired by multi-instance learning \cite{maron1998framework}, Riedel et al. \shortcite{riedel2010modeling} relaxed the strong assumption and replaced {\it all sentences} with {\it at least one sentence}. Hoffmann et al. \shortcite{hoffmann-EtAl:2011:ACL-HLT2011} pointed out that many entity pairs have more than one relation. They extended the multi-instance learning framework \cite{riedel2010modeling} to the multi-label circumstance. Surdeanu et al. \shortcite{Surdeanu2012} proposed a novel approach to multi-instance multi-label learning for relation extraction, which jointly modeled all the sentences in texts and all labels in knowledge bases for a given entity pair.
Other literatures
\cite{Takamatsu2012, min-EtAl:2013:NAACL-HLT, zhang-EtAl:2013:Short3, xu-EtAl:2013:Short2} addressed more specific issues, like how to construct the negative class in learning or how to adopt more information, such as name entity tags, to improve the performance.

Our work is more relevant to Riedel et al.'s \shortcite{riedel-EtAl:2013:NAACL-HLT} which considered the task as a matrix factorization problem. Their approach is composed of several models, such as PCA \cite{Collins2001} and collaborative filtering \cite{Koren2008}. However, they did not concern about the data noise brought by the basic assumption of distant supervision.

\section{Model}

We apply a new technique in the field of applied mathematics, i.e., low-rank matrix completion with convex optimization. The breakthrough work on this topic was made by Cand{\`e}s and Recht \shortcite{candes2009exact} who proved that most low-rank matrices can be perfectly recovered from an incomplete set of entries. This promising theory has been successfully applied on many active research areas, such as computer vision \cite{Cabral2011}, recommender system \cite{Rennie2005} and system controlling \cite{Fazel2001}.
Our models for relation extraction are based on the theoretic framework proposed by Goldberg et al. \shortcite{Goldberg2010}, which formulated the multi-label transductive learning as a matrix completion problem. The new framework for classification enhances the robustness to data noise by penalizing different cost functions for features and labels.

\subsection{Formulation}
Suppose that we have built a training corpus for relation classification with $n$ items (entity pairs), $d$-dimensional textual features, and $t$ labels (relations), based on the basic alignment assumption proposed by Mintz et al. \shortcite{mintz2009distant}. Let $X_{train} \in \mathbb{R}^{n \times d}$ and $Y_{train} \in \mathbb{R}^{n \times t}$ denote the feature matrix and the label matrix for training, respectively. The linear classifier we adopt aims to explicitly learn the weight matrix ${\bf W} \in \mathbb{R}^{d \times t}$ and the bias column vector ${\bf b} \in \mathbb{R}^{t \times 1}$ with the constraint of minimizing the loss function $l$,
\vspace{-5pt}
\begin{equation}
\mathop{\argmin}_{{\bf W, b}} ~~ l(Y_{train}, \left[ {\begin{array}{*{20}{c}}
{\bf 1}&{{X_{train}}}
\end{array}} \right]\left[ {\begin{array}{*{20}{c}}
{\bf b}^T\\
{\bf W}
\end{array}} \right]),
\end{equation}
where {\bf 1} is the all-one column vector.
Then we can predict the label matrix $Y_{test} \in \mathbb{R} ^ {m \times t}$ of $m$ testing items with respect to the feature matrix $X_{test} \in \mathbb{R} ^ {m \times d}$.
Let
\vspace{-5pt}
\[
{\bf Z} = \left[\begin{array}{*{20}{c}}
{{X_{train}}}&{{Y_{train}}}\\
{{X_{test}}}&{{Y_{test}}}
\end{array}\right].
\]
This linear classification problem can be transformed into completing the unobservable entries in $Y_{test}$ by means of the observable entries in $X_{train}$, $Y_{train}$ and $X_{test}$,
based on the assumption that the rank of matrix ${\bf Z} \in \mathbb{R}^{(n + m) \times (d + t)}$ is low. The model can be written as,
\vspace{-5pt}
\begin{equation}
\begin{split}
&\mathop{\argmin}_{{\bf Z} \in \mathbb{R}^{(n + m) \times (d + t)}} ~~ \text {rank}({\bf Z}) \\
&\text{s.t.} ~~\forall(i, j) \in \Omega_X, ~~z_{ij} = x_{ij}, \\
&~~~~~~~~ (1 \leq i \leq n + m, ~~ 1 \leq j \leq d),\\
&~~~~~~~~\forall(i, j) \in \Omega_Y, ~~z_{i(j+d)} = y_{ij},\\
&~~~~~~~~ (1 \leq i \leq n, ~~ 1 \leq j \leq t),
\end{split}
\end{equation}
 where we use $\Omega_X$ to represent the index set of observable feature entries in $X_{train}$ and $X_{test}$, and $\Omega_Y$ to denote the index set of observable label entries in $Y_{train}$.

Formula (2) is usually impractical for real problems as the entries in the matrix ${\bf Z}$ are corrupted by noise. We thus define
\vspace{-5pt}
\[
{\bf Z} = {\bf Z^*} + {\bf E},
\]
where ${\bf Z^*}$ as the underlying low-rank matrix
\vspace{-5pt}
\[
{\bf Z^{*}} = \left[ {\begin{array}{*{20}{c}}
{{X^*}}&{{Y^*}}
\end{array}} \right] =
\left[ {\begin{array}{*{20}{c}}
{X_{train}^*}&{Y_{train}^*}\\
{X_{test}^*}&{Y_{test}^*}
\end{array}} \right],
\]
and {\bf E} is the error matrix
\vspace{-5pt}
\[
{\bf E} =
\left[ {\begin{array}{*{20}{c}}
{E_{X_{train}}}&{E_{Y_{train}}}\\
{E_{X_{test}}}&{0}
\end{array}} \right].
\]
The rank function in Formula (2) is a non-convex function that is difficult to be optimized. The surrogate of the function can be the convex nuclear norm ${\bf ||Z||_*} = \sum \sigma_k({\bf Z})$ \cite{candes2009exact}, where $\sigma_k$ is the $k$-$th$ largest singular value of ${\bf Z}$. To tolerate the noise entries in the error matrix ${\bf E}$, we minimize the cost functions $C_x$ and $C_y$ for features and labels respectively, rather than using the hard constraints in Formula (2).

According to Formula (1), ${\bf Z^*} \in \mathbb{R} ^ {(n + m) \times (d + t)}$ can be represented as $[X^*, {\bf W}X^*]$ instead of $[X^*, Y^*]$, by explicitly modeling the bias vector ${\bf b}$. Therefore, this convex optimization model is called {\bf DRMC-b},
\vspace{-5pt}
\begin{equation}
\begin{split}
&\mathop{\argmin}_{{\bf Z, b}} ~~\mu||{\bf Z}||_* + \frac{1}{|{\Omega_X}|}\sum_{(i, j) \in {\Omega_X}}{C_x(z_{ij}, x_{ij})} \\
& ~~ ~~ ~~ ~~ ~~ ~~ ~~ + \frac{\lambda }{{|{\Omega_Y}|}}\sum_{(i, j) \in {\Omega_Y}}{C_y(z_{i(j+d)} + b_j, y_{ij})}, \\
\end{split}
\end{equation}
where $\mu$ and $\lambda$ are the positive trade-off weights. More specifically, we minimize the nuclear norm $||{\bf Z}||_*$ via employing the regularization terms, i.e., the cost functions $C_x$ and $C_y$ for features and labels.

If we implicitly model the bias vector ${\bf b}$, ${\bf Z^*} \in \mathbb{R} ^ {(n + m) \times (1 + d + t)}$ can be denoted by $[{\bf 1}, X^*, {\bf W^{'}}X^*]$ instead of $[X^*, Y^*]$, in which ${\bf W^{'}}$ takes the role of $[{\bf b}^T; {\bf W}]$ in DRMC-b. Then we derive another optimization model called {\bf DRMC-1},
\vspace{-5pt}
\begin{equation}
\begin{split}
&\mathop{\argmin}_{{\bf Z}} ~~\mu||{\bf Z}||_* + \frac{1}{|{\Omega_X}|}\sum_{(i, j) \in {\Omega_X}}{C_x(z_{i(j+1)}, x_{ij})}\\
&~~ ~~ ~~ ~~ ~~ ~~ ~~ ~~ ~~ ~~ ~~  + \frac{\lambda }{{|{\Omega_Y}|}}\sum_{(i, j) \in {\Omega_Y}}{C_y(z_{i(j + d + 1)}, y_{ij})}  \\
& ~~~~~~~\text{s.t.} ~~~~~~~~~~~~~~~~~~{\bf Z}(:, 1) = {\bf 1}, \\
\end{split}
\end{equation}
where ${\bf Z}(:, 1)$ denotes the first column of ${\bf Z}$.

For our relation classification task, both features and labels are binary. We assume that the actual entry $u$ belonging to the underlying matrix ${\bf Z^*}$ is randomly generated via a sigmoid function \cite{jordan1995logistic}: $Pr(u|v) = 1 / (1 + e^{-uv})$, given the observed binary entry $v$ from the observed sparse matrix ${\bf Z}$. Then, we can apply the log-likelihood cost function to measure the conditional probability and derive the {\it logistic cost function} for $C_x$ and $C_y$,
\vspace{-5pt}
\[
C(u,v) = -\log Pr(u|v) = \log(1 + e^{-uv}),
\]

After completing the entries in $Y_{test}$, we adopt the sigmoid function to calculate the conditional probability of relation $r_j$, given entity pair $p_i$ pertaining to $y_{ij}$ in $Y_{test}$,
\vspace{-5pt}
\[
Pr(r_j|p_i) = \frac{1} { 1 + e^{-y_{ij}}}, ~~~ y_{ij} \in Y_{test}.
\]
Finally, we can achieve Top-N predicted relation instances via ranking the values of $Pr(r_j|p_i)$.
\section{Algorithm}
The matrix rank minimization problem is NP-hard. Therefore, Cand{\'e}s and Recht \shortcite{candes2009exact} suggested to use a convex relaxation, the nuclear norm minimization instead. Then, Ma et al. \shortcite{Ma2011} proposed the fixed point continuation (FPC) algorithm which is fast and robust. Moreover, Goldfrab and Ma \shortcite{goldfarb2011convergence} proved the convergence of the FPC algorithm for solving the nuclear norm minimization problem. We thus adopt and modify the algorithm aiming to find the optima for our noise-tolerant models, i.e., Formulae (3) and (4).

\subsection{Fixed point continuation for DRMC-b}
Algorithm 1 describes the modified FPC algorithm for solving DRMC-b, which contains two steps for each iteration,

{\bf Gradient step:} In this step, we infer the matrix gradient $g({\bf Z})$ and bias vector gradient $g({\bf b})$ as follows,
\vspace{-5pt}
\[
g(z_{ij})=
\begin{cases}
\frac{1}{|\Omega_X|} \frac{-x_{ij}}{1 + e^{x_{ij}z_{ij}}}, & (i,j) \in \Omega_X\\
\frac{\lambda}{|\Omega_Y|} \frac{-y_{i(j-d)}}{1 + e^{y_{i(j-d)}(z_{ij} + b_j)}}, & (i,j-d) \in \Omega_Y\\
0,& otherwise
\end{cases}
\]
and
\vspace{-5pt}
\[
g(b_j) = \frac{\lambda}{|\Omega_Y|} \sum_{i:(i, j) \in \Omega_Y}{\frac{-y_{ij}}{1 + e^{y_{ij}(z_{i(j+d)} + b_j)}}}.
\]
We use the gradient descents ${\bf A} = {\bf Z} - \tau_zg({\bf Z})$ and ${\bf b} = {\bf b} - \tau_b g({\bf b})$ to gradually find the global minima of the cost function terms in Formula (3), where $\tau_z$ and $\tau_b$ are step sizes.

{\bf Shrinkage step:} The goal of this step is to minimize the nuclear norm ${\bf ||Z||_*}$ in Formula (3). We perform the singular value decomposition (SVD) \cite{golub1965calculating} for ${\bf A}$ at first, and then cut down each singular value. During the iteration, any negative value in ${\bf \Sigma - \tau_z \mu}$ is assigned by zero, so that the rank of reconstructed matrix ${\bf Z} $ will be reduced, where ${\bf Z} = {\bf U} max({\bf \Sigma - \tau_z \mu}, 0) {\bf V^T}$.

\begin{algorithm}
\caption{FPC algorithm for solving DRMC-b}
\label{alg:DRMC-b}
\begin{algorithmic}
\REQUIRE ~~ \\Initial matrix ${\bf Z_0} $, bias ${\bf b_0}$;  
   Parameters $\mu, \lambda$;\\
   Step sizes $\tau_z, \tau_b$.\\
   \vspace{-5pt}
\noindent\rule{0.43\textwidth}{0.4pt}\\
Set ${\bf Z} = {\bf Z_0}$, ${\bf b} = {\bf b_0}.$
\FOR {$\mu = \mu_1 > \mu_2 > ... > \mu_F$}
\WHILE {relative error $>\varepsilon$}
    \STATE Gradient step: \\${\bf A} = {\bf Z} - \tau_zg({\bf Z}), {\bf b} = {\bf b} - \tau_b g({\bf b}).$
    \STATE Shrinkage step:\\
    ${\bf U}{\bf \Sigma}{\bf V^T} = \text{SVD}({\bf A})$,\\
    $ {\bf Z} = {\bf U} ~ max({\bf \Sigma - \tau_z \mu}, 0) ~ {\bf V^T}.$
\ENDWHILE

\ENDFOR
\\
\vspace{-5pt}
\noindent\rule{0.43\textwidth}{0.4pt}\\
\ENSURE ~~  Completed Matrix {\bf Z}, bias {\bf b}.   
\end{algorithmic}
\end{algorithm}
To accelerate the convergence, we use a continuation method to improve the speed. $\mu$ is initialized by a large value $\mu_1$, thus resulting in the fast reduction of the rank at first. Then the convergence slows down as $\mu$ decreases while obeying $\mu_{k+1} = max(\mu_k\eta_{\mu}, \mu_F)$. $\mu_F$ is the final value of $\mu$, and $\eta_{\mu}$ is the decay parameter.

For the stopping criteria in inner iterations, we define the {\it relative error} to measure the residual of matrix ${\bf Z}$ between two successive iterations,
\vspace{-5pt}
\[
\frac{||{\bf Z}^{k+1} - {\bf Z}^k||_F}{max(1, ||{\bf Z}^k||_F)} \leq \varepsilon,
\]
where $\varepsilon$ is the convergence threshold.

\subsection{Fixed point continuation for DRMC-1}
Algorithm 2 is similar to Algorithm 1 except for two differences. First, there is no bias vector {\bf b}. Second, a projection step is added to enforce the first column of matrix {\bf Z} to be {\bf 1}. In addition, The matrix gradient $g({\bf Z})$ for DRMC-1 is
\vspace{-5pt}
\[
g(z_{ij})=
\begin{cases}
\frac{1}{|\Omega_X|} \frac{-x_{i(j-1)}}{1 + e^{x_{i(j-1)}z_{ij}}}, & (i,j-1) \in \Omega_X\\
\frac{\lambda}{|\Omega_Y|} \frac{-y_{i(j-d-1)}}{1 + e^{y_{i(j-d-1)}z_{ij}}}, & (i,j-d-1) \in \Omega_Y\\
0,& otherwise
\end{cases}.
\]

\begin{algorithm}
\caption{FPC algorithm for solving DRMC-1}
\label{alg:DRMC-1}
\begin{algorithmic}
\REQUIRE ~~ \\Initial matrix ${\bf Z_0} $; 
   Parameters $\mu, \lambda$;\\
   Step sizes $\tau_z$.\\
\vspace{-5pt}
\noindent\rule{0.43\textwidth}{0.4pt}\\
Set ${\bf Z} = {\bf Z_0}.$
\FOR {$\mu = \mu_1 > \mu_2 > ... > \mu_F$}
\WHILE {relative error $>\varepsilon$}
    \STATE Gradient step: ${\bf A} = {\bf Z} - \tau_zg({\bf Z}).$
    \STATE Shrinkage step:
    \\${\bf U}{\bf \Sigma}{\bf V^T} = \text {SVD}({\bf A})$,\\
    $ {\bf Z} = {\bf U}~ max({\bf \Sigma - \tau_z \mu}, 0) ~{\bf V^T}.$
    \STATE Projection step:
    ${\bf Z}(:, 1) = {\bf 1}.$
\ENDWHILE

\ENDFOR
\\
\vspace{-5pt}
\noindent\rule{0.43\textwidth}{0.4pt}\\
\ENSURE ~~  Completed Matrix {\bf Z}. 
\end{algorithmic}
\end{algorithm}

\section{Experiments}
\begin{table*}
\centering
\begin{tabular}{|C{1.7cm}|C{2cm}|C{2cm}|C{2.5cm}|C{2cm}|C{2cm}|}
  \hline
      Dataset &\# of training tuples & \# of testing tuples &  \% with more than one label & \# of features & \# of relation labels \tabularnewline
                \hline
         NYT'10 & 4,700 & 1,950 & 7.5\% & 244,903 & 51 \tabularnewline
        NYT'13& 8,077 & 3,716 & 0\% & 1,957 &51 \tabularnewline
  \hline
\end{tabular}

\caption{Statistics about the two widely used datasets.}

\end{table*}

\begin{table*}
\centering
\begin{tabular}{|c|c|c|c|c|c|}
  \hline
        Model & NYT'10 ($\theta$=2) &  NYT'10 ($\theta$=3) &  NYT'10 ($\theta$=4) &  NYT'10 ($\theta$=5) &  NYT'13\\
          \hline
             DRMC-b & 51.4 $\pm$ 8.7 (51) & 45.6 $\pm$ 3.4 (46) & 41.6 $\pm$ 2.5 (43) & 36.2 $\pm$ 8.8(37) & 84.6 $\pm$ 19.0 (85)\\
  DRMC-1 & 16.0 $\pm$ 1.0 (16) &  16.4 $\pm$ 1.1(17) & 16 $\pm$ 1.4 (17) & 16.8 $\pm$ 1.5(17) &  15.8 $\pm$ 1.6 (16) \\
  \hline
\end{tabular}

\caption{The range of optimal ranks for DRMC-b and DRMC-1 through five-fold cross validation. The threshold $\theta$ means filtering the features that appear less than $\theta$ times.
The values in brackets pertaining to DRMC-b and DRMC-1 are the exact optimal ranks that we choose for the completed matrices on testing sets.}
\vspace{-5pt}
\end{table*}
In order to conduct reliable experiments, we adjust and estimate the parameters for our approaches, DRMC-b and DRMC-1, and compare them with other four kinds of landmark methods \cite{mintz2009distant, hoffmann-EtAl:2011:ACL-HLT2011, Surdeanu2012, riedel-EtAl:2013:NAACL-HLT} on two public datasets.
\subsection{Dataset}

The two widely used datasets that we adopt are both automatically generated by aligning Freebase to New York Times corpora.
The first dataset\footnote{http://iesl.cs.umass.edu/riedel/ecml/}, NYT'10, was developed by Riedel et al. \shortcite{riedel2010modeling}, and also used by Hoffmann et al. \shortcite{hoffmann-EtAl:2011:ACL-HLT2011} and Surdeanu et al. \shortcite{Surdeanu2012}. Three kinds of features, namely, lexical, syntactic and named entity tag features, were extracted from relation mentions.
The second dataset\footnote{http://iesl.cs.umass.edu/riedel/data-univSchema/}, NYT'13, was also released by Riedel et al. \shortcite{riedel-EtAl:2013:NAACL-HLT}, in which they only regarded the lexicalized dependency path between two entities as features.
Table 1 shows that the two datasets differ in some main attributes. More specifically, NYT'10 contains much higher dimensional features than NYT'13, whereas fewer training and testing items.
\begin{figure*}[htb]
  \centering
  \subfigure[DRMC-b on NYT'10 validation set ($\theta = 5$).]{
    \raisebox{-1cm}
    {\includegraphics[width=0.5\textwidth]{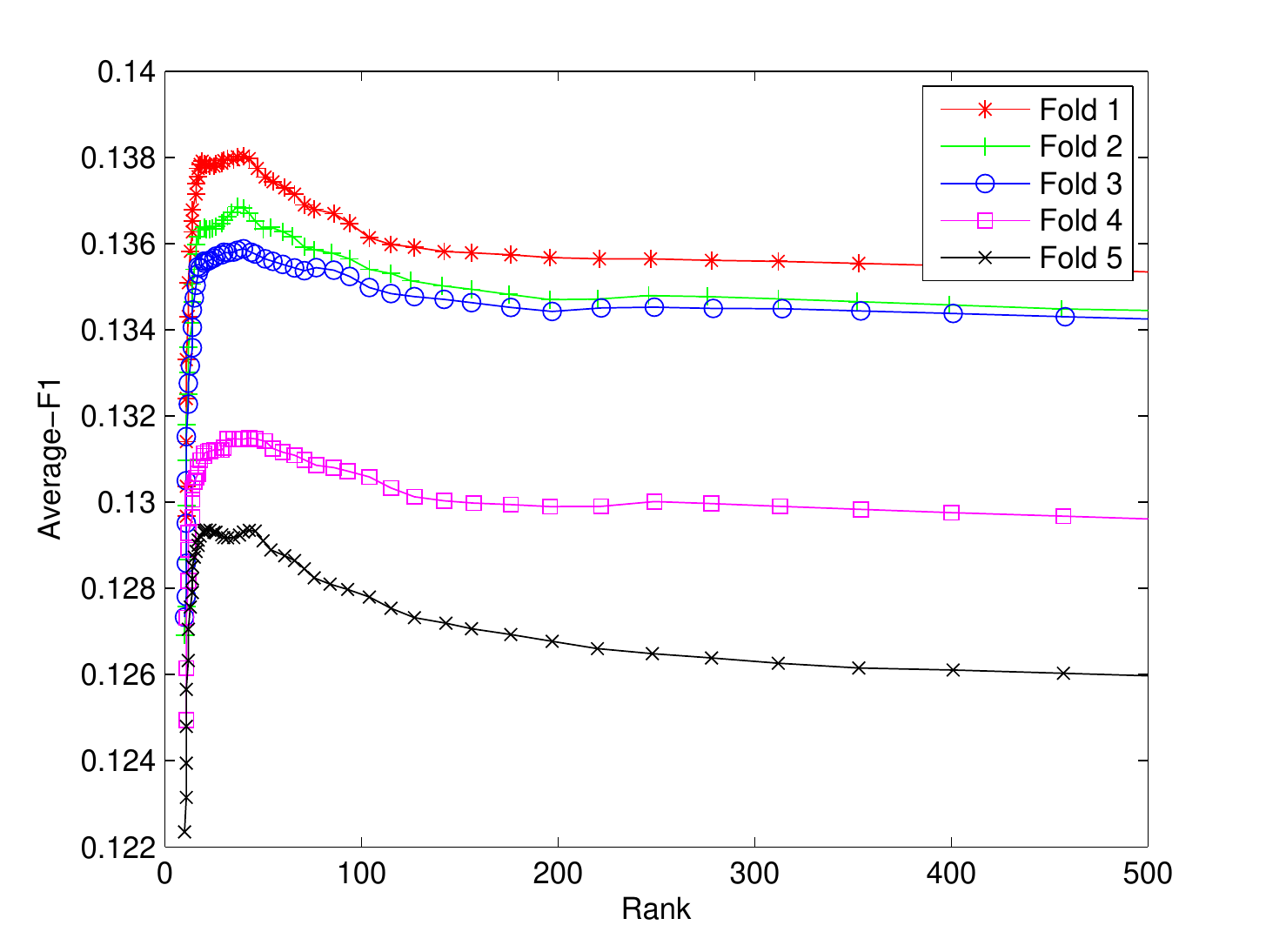}}
    }
   \vspace{-5pt}
    \hspace{-23pt}
      \subfigure[DRMC-1 on NYT'10 validation set ($\theta = 5$).]{
    \raisebox{-1cm}{
    \includegraphics[width=0.5\textwidth]{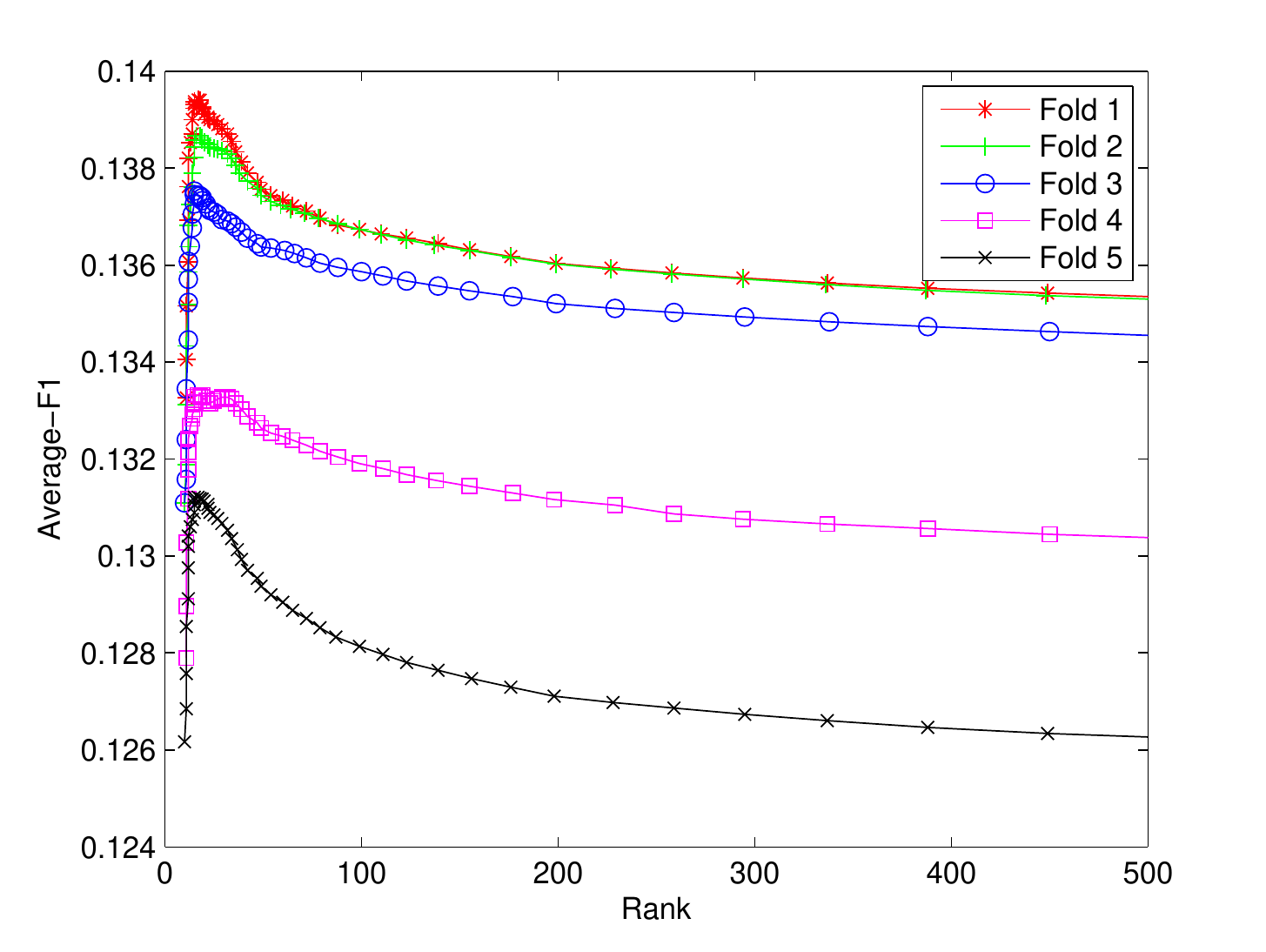}}
    }

    \subfigure[DRMC-b on NYT'13 validation set.]{
        \raisebox{-1cm}{
    \includegraphics[width=0.5\textwidth]{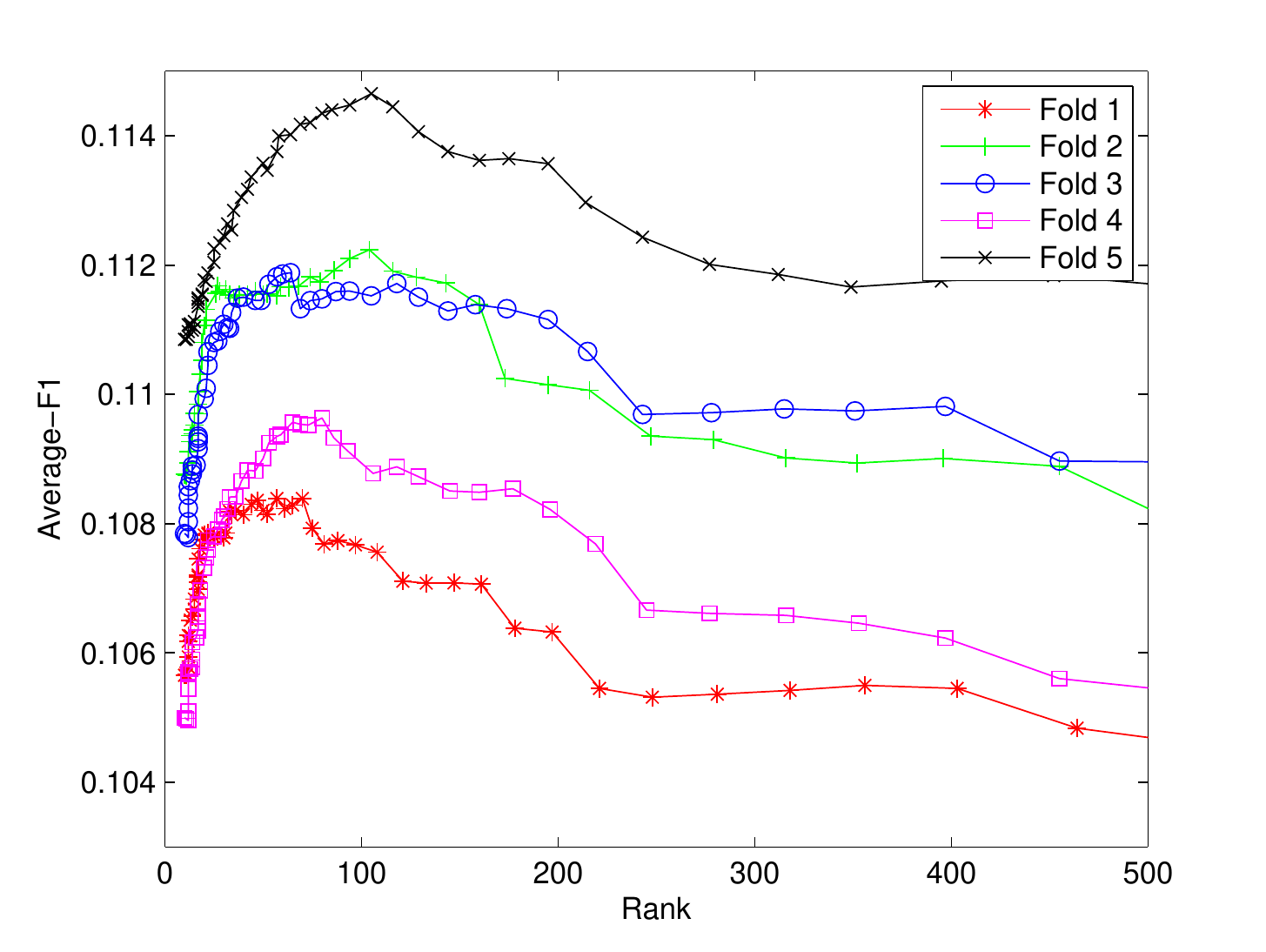}}

    }
    \hspace{-23pt}
    \subfigure[DRMC-1 on NYT'13 validation set.]{
        \raisebox{-1cm}{
    \includegraphics[width=0.5\textwidth]{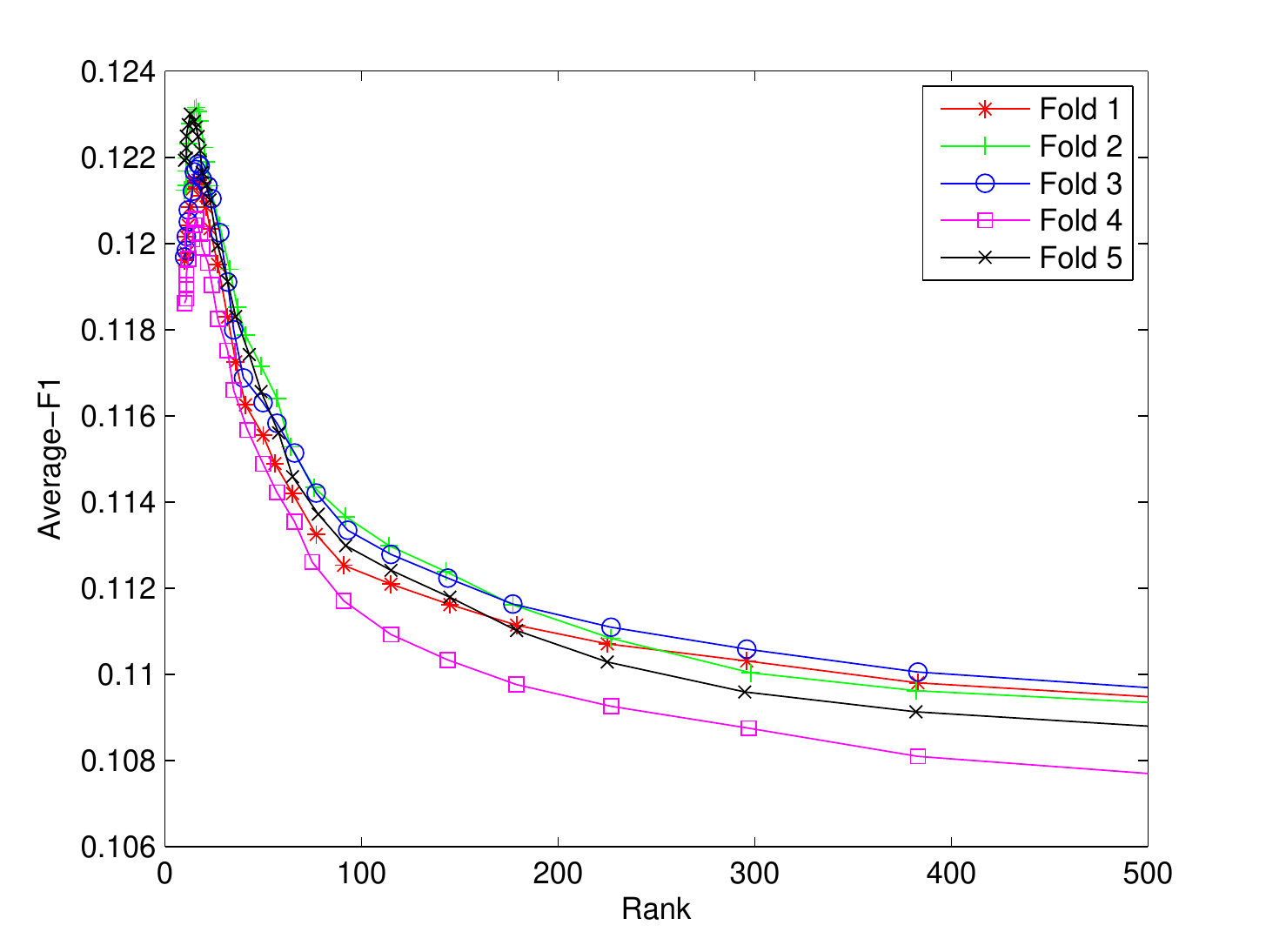}
    }}

  \caption{Five-fold cross validation for rank estimation on two datasets.}

\end{figure*}

\subsection{Parameter setting}
In this part, we address the issue of setting parameters: the trade-off weights $\mu$ and $\lambda$, the step sizes $\tau_z$ and $\tau_b$, and the decay parameter $\eta_{\mu}$.

We set $\lambda = 1$ to make the contribution of the cost function terms for feature and label matrices equal in Formulae (3) and (4).
$\mu$ is assigned by a series of values obeying $\mu_{k+1} = max(\mu_k\eta_{\mu}, \mu_F)$.
We follow the suggestion in \cite{Goldberg2010} that $\mu$ starts at $\sigma_1\eta_{\mu}$, and $\sigma_1$ is the largest singular value of the matrix {\bf Z}. We set $\eta_{\mu} = 0.01$. The final value of $\mu$, namely $\mu_F$, is equal to $0.01$.
Ma et al. \shortcite{Ma2011} revealed that as long as the non-negative step sizes satisfy $\tau_z < min(\frac{4|\Omega_Y|}{\lambda}, |\Omega_X|)$ and $\tau_b < \frac{4|\Omega_Y|}{\lambda(n+m)}$, the FPC algorithm will guarantee to converge to a global optimum. Therefore, we set $\tau_z = \tau_b = 0.5$ to satisfy the above constraints on both two datasets.

\subsection{Rank estimation}

Even though the FPC algorithm converges in iterative fashion, the value of $\varepsilon$ varying with different datasets is difficult to be decided. In practice, we record the rank of matrix {\bf Z} at each round of iteration until it converges at a rather small threshold $\varepsilon = 10^{-4}$. The reason is that we suppose the optimal low-rank representation of the matrix {\bf Z} conveys the truly effective information about underlying semantic correlation between the features and the corresponding labels.

We use the five-fold cross validation on the validation set and evaluate the performance on each fold with different ranks. At each round of iteration, we gain a recovered matrix and average the F1\footnote{$F1 = \frac{2 \times precision \times recall}{precision + recall}$} scores from Top-5 to Top-all predicted relation instances to measure the performance. Figure 3 illustrates the curves of average F1 scores. After recording the rank associated with the highest F1 score on each fold, we compute the mean and the standard deviation to estimate the range of optimal rank for testing. Table 2 lists the range of optimal ranks for DRMC-b and DRMC-1 on NYT'10 and NYT'13.

On both two datasets, we observe an identical phenomenon that the performance gradually increases as the rank of the matrix declines before reaching the optimum. However, it sharply decreases if we continue reducing the optimal rank. An intuitive explanation is that the high-rank matrix contains much noise and the model tends to be overfitting, whereas the matrix of excessively low rank is more likely to lose principal information and the model tends to be underfitting.

\subsection{Method Comparison}
\begin{figure*}[htb]
  \centering
  \subfigure[NYT'10 testing set ($\theta = 5$).]{
  \raisebox{-1cm}{
    \includegraphics[width=0.5\textwidth]{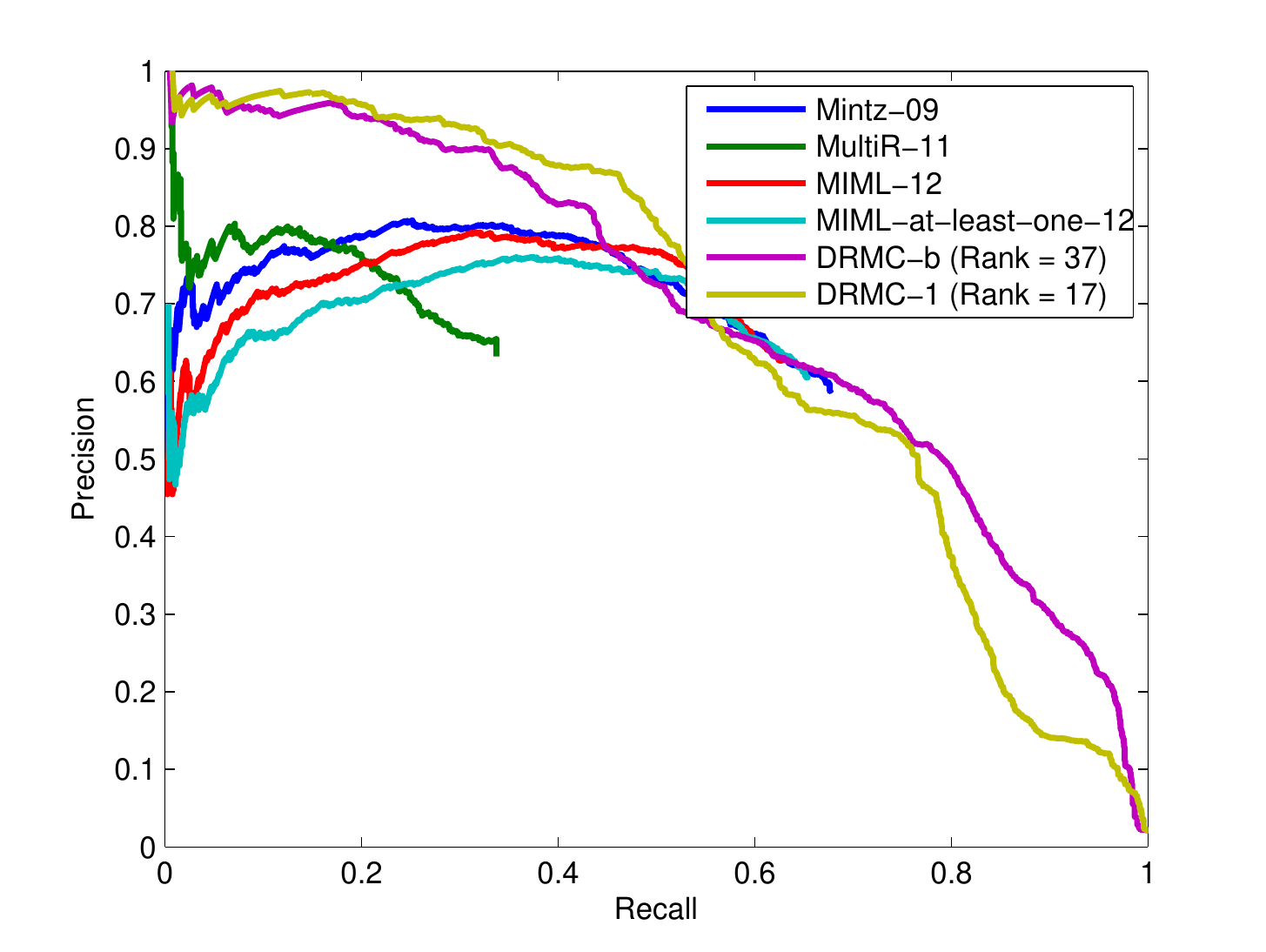}}
    }
    \hspace{-23pt}
    \subfigure[NYT'13 testing set.]{
    \raisebox{-1cm}{
    \includegraphics[width=0.5\textwidth]{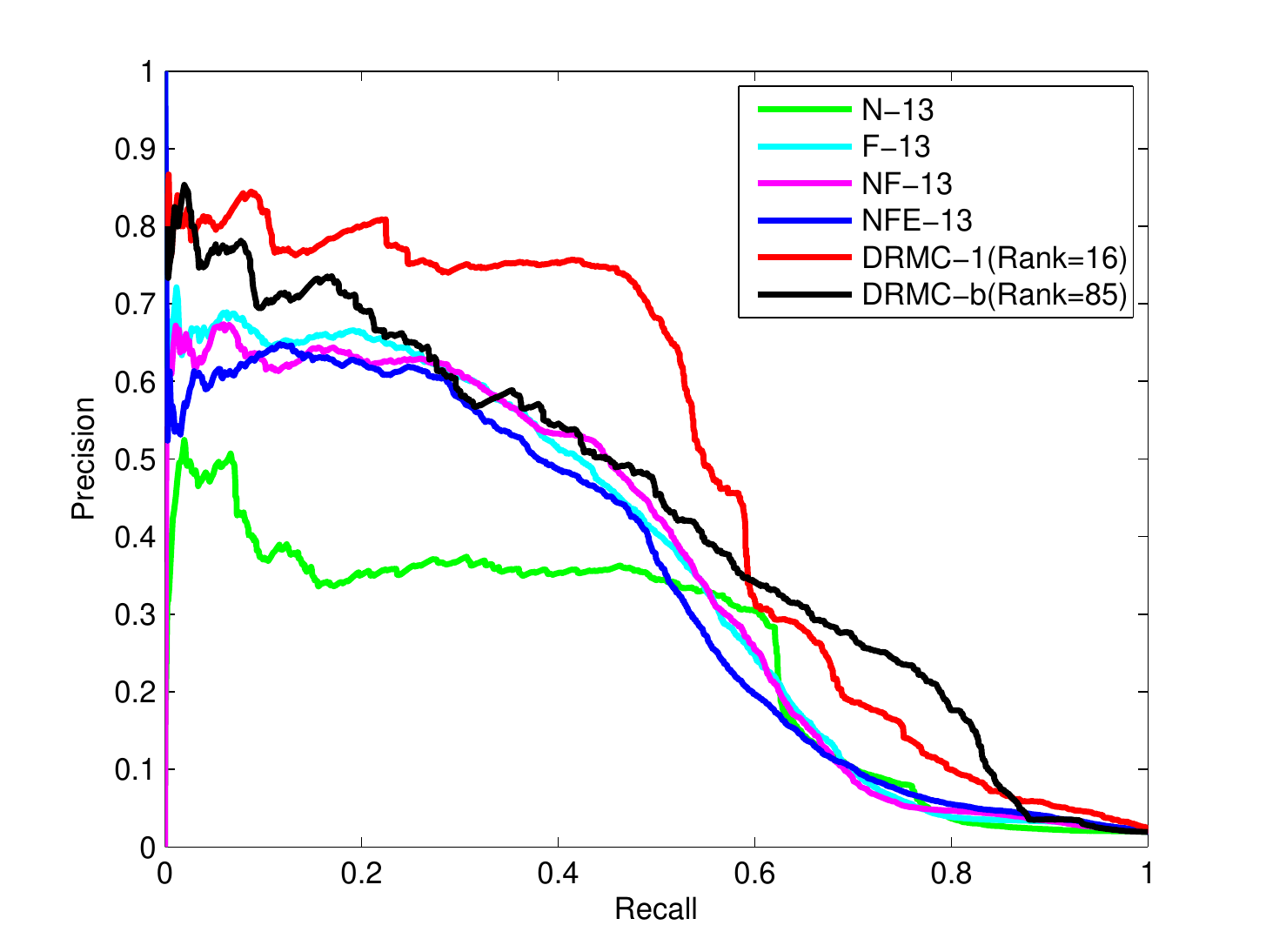}}
    }

  \caption{Method comparison on two testing sets.}

\end{figure*}
\begin{figure*}[htb]
  \centering
  \subfigure[NYT'10 testing set ($\theta = 5$).]{
  \raisebox{-1cm}{
    \includegraphics[width=0.5\textwidth]{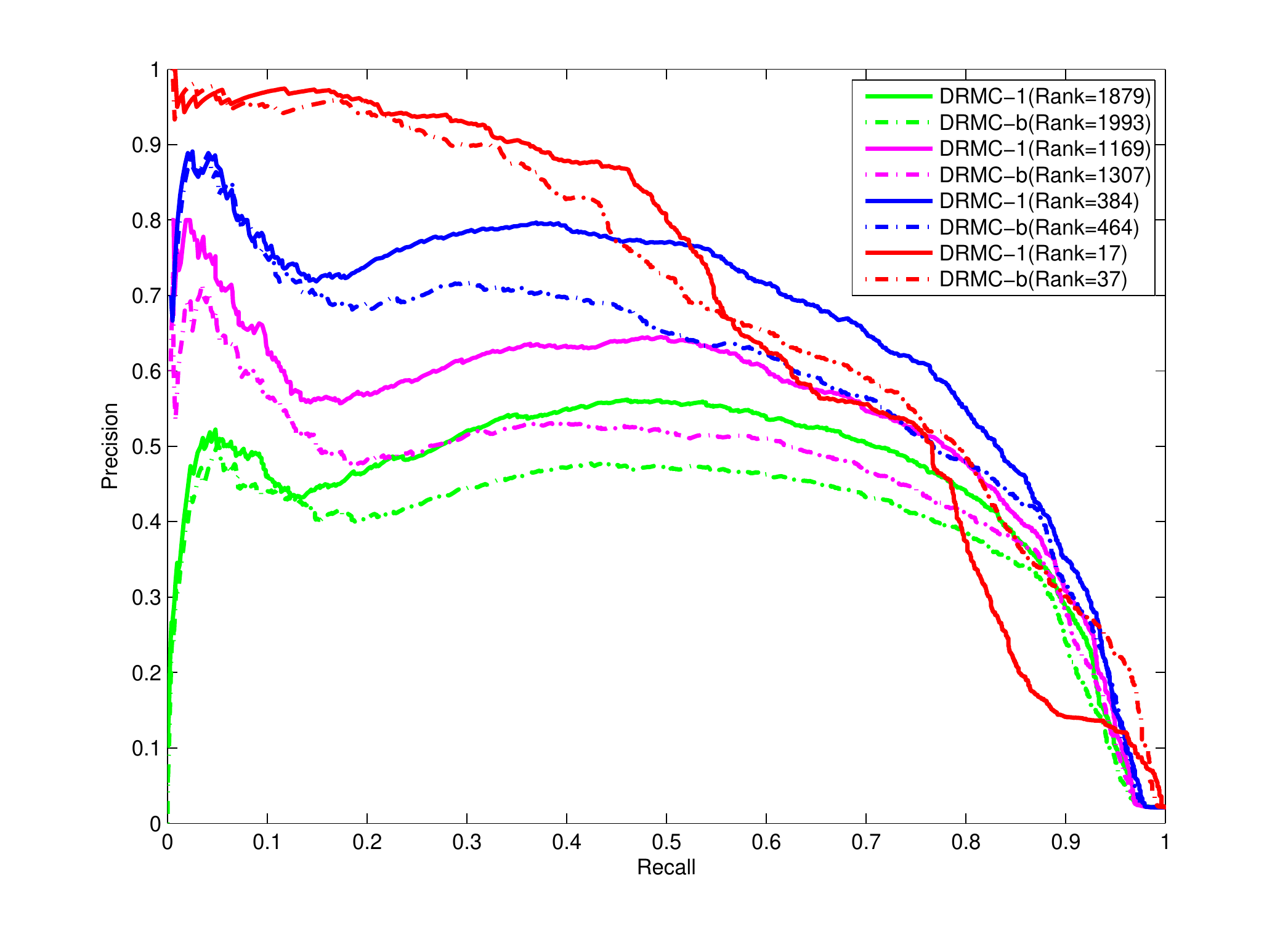}}
    }
    \hspace{-23pt}
      \subfigure[NYT'13 testing set.]{
      \raisebox{-1cm}{
    \includegraphics[width=0.5\textwidth]{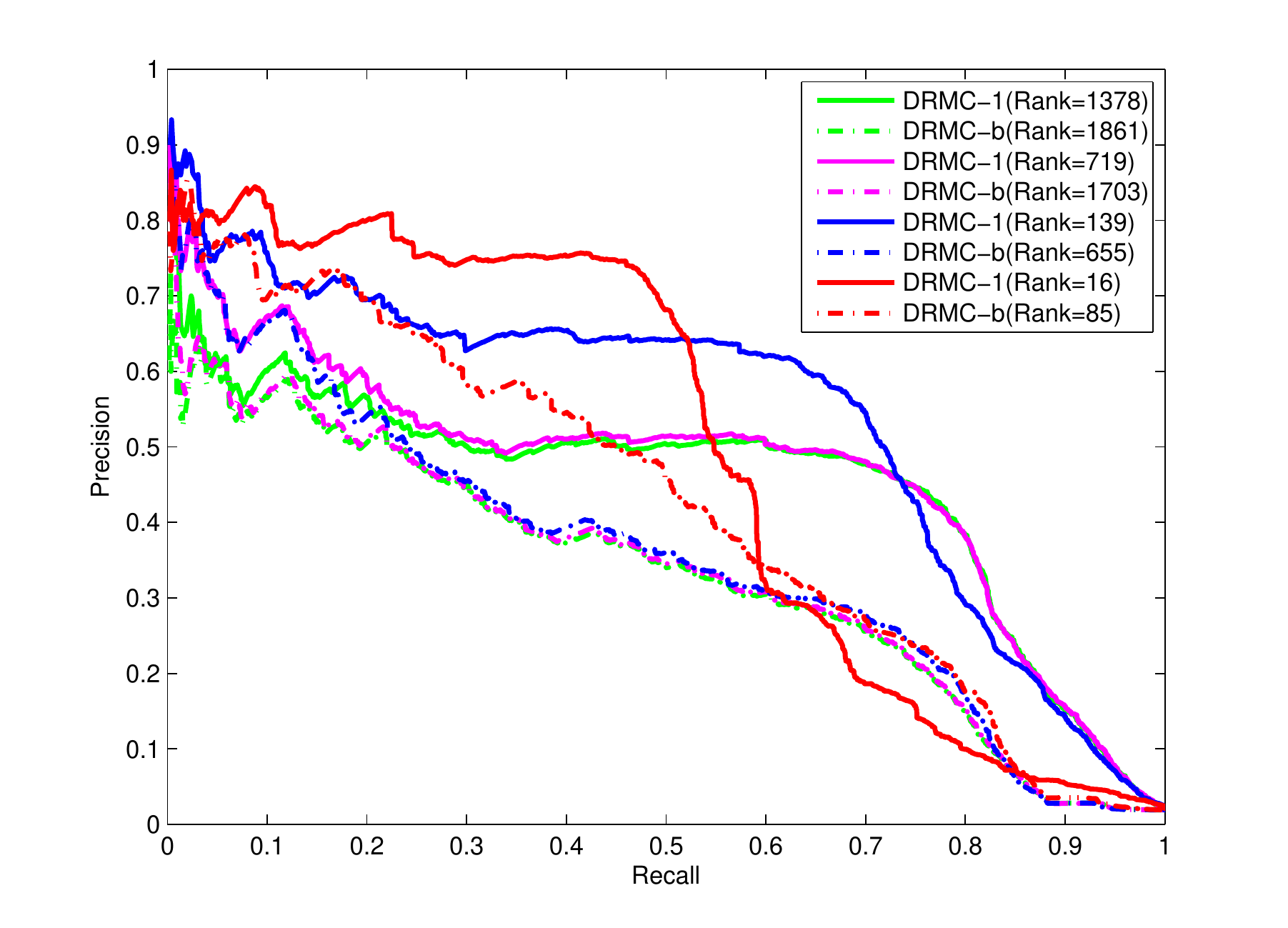}
    }}

  \caption{Precision-Recall curve for DRMC-b and DRMC-1 with different ranks on two testing sets.}

\end{figure*}


Firstly, we conduct experiments to compare our approaches with Mintz-09 \cite{mintz2009distant}, MultiR-11 \cite{hoffmann-EtAl:2011:ACL-HLT2011}, MIML-12 and MIML-at-least-one-12 \cite{Surdeanu2012} on NYT'10 dataset. Surdeanu et al. \shortcite{Surdeanu2012} released the open source code\footnote{http://nlp.stanford.edu/software/mimlre.shtml} to reproduce the experimental results on those previous methods. Moreover, their programs can control the feature sparsity degree through a threshold $\theta$ which filters the features that appears less than $\theta$ times. They set $\theta = 5$ in the original code by default. Therefore, we follow their settings and adopt the same way to filter the features. In this way, we guarantee the fair comparison for all methods. Figure 4 (a) shows that our approaches achieve the significant improvement on performance.

We also perform the experiments to compare our approaches with the state-of-the-art NFE-13\footnote{Readers may refer to the website, http://www.riedelcastro.org/uschema for the details of those methods. We bypass the description due to the limitation of space.} \cite{riedel-EtAl:2013:NAACL-HLT} and its sub-methods (N-13, F-13 and NF-13) on NYT'13 dataset. Figure 4 (b) illustrates that our approaches still outperform the state-of-the-art methods.
\begin{figure*}[htb]
  \centering
  \subfigure{
    \includegraphics[width=0.26\textwidth]{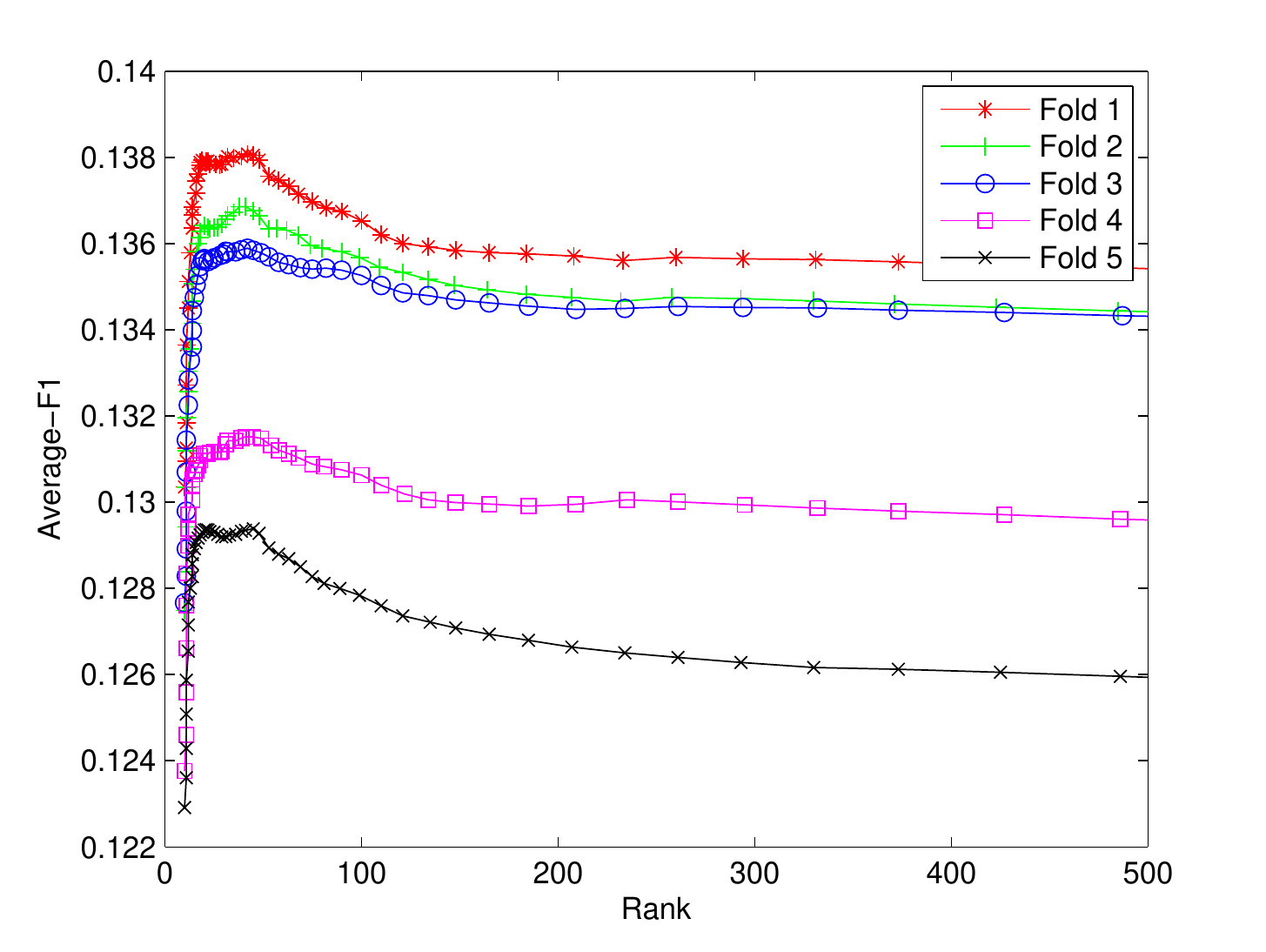}}
    \hspace{-15pt}
  \subfigure{
    \includegraphics[width=0.26\textwidth]{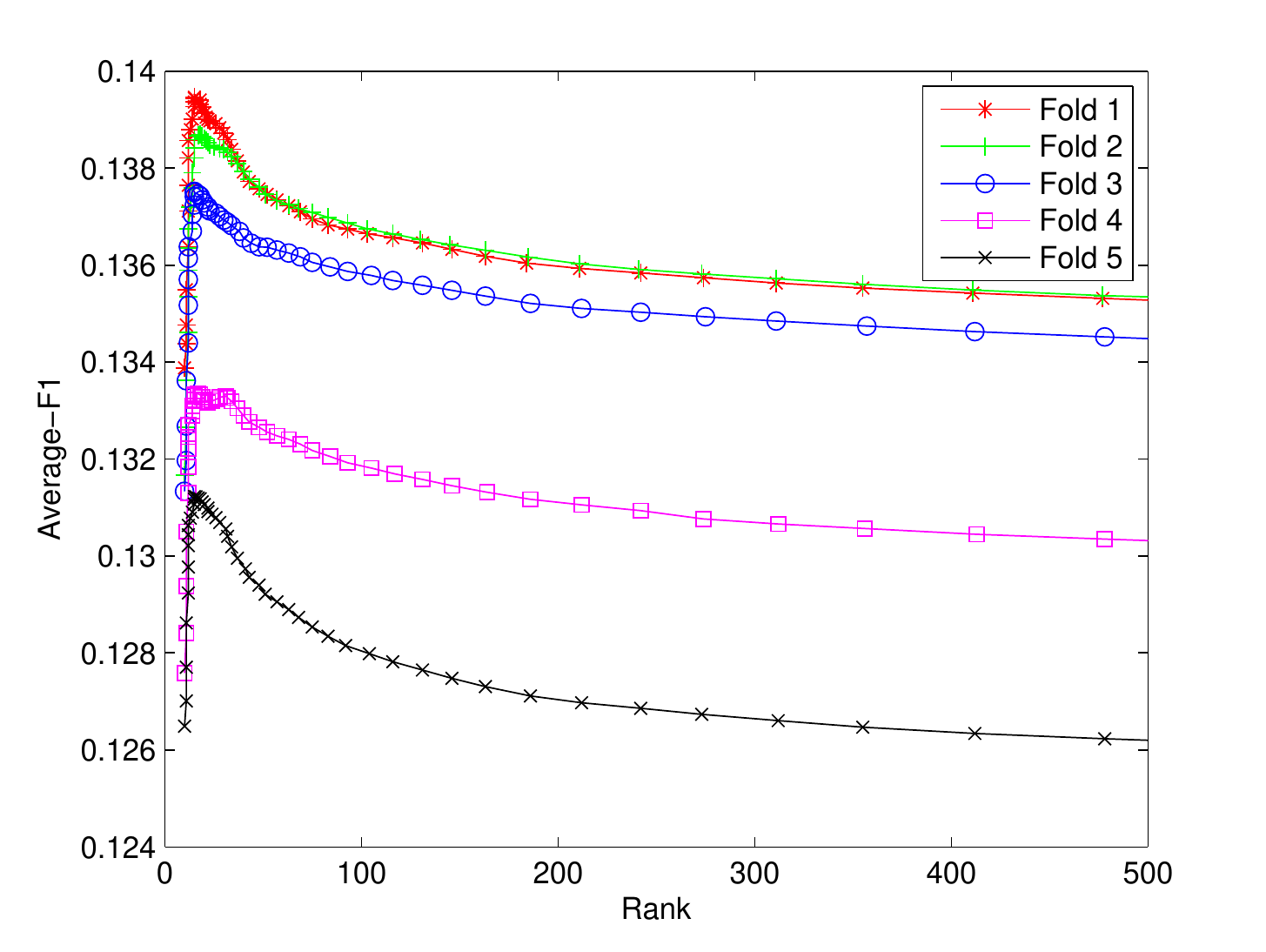}}
    \hspace{-15pt}
  \subfigure{
    \includegraphics[width=0.26\textwidth]{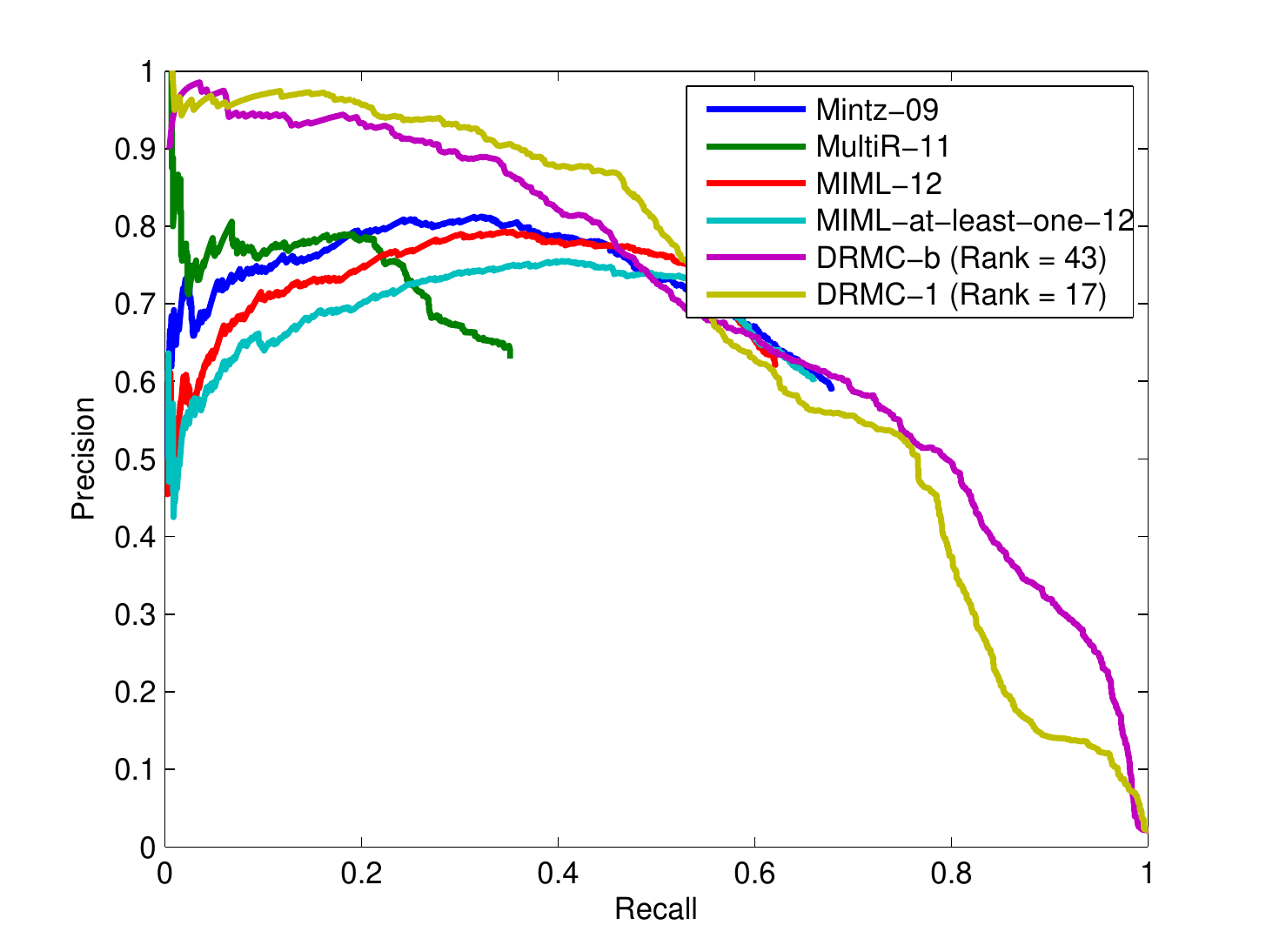}}
    \hspace{-15pt}
  \subfigure{
    \includegraphics[width=0.26\textwidth]{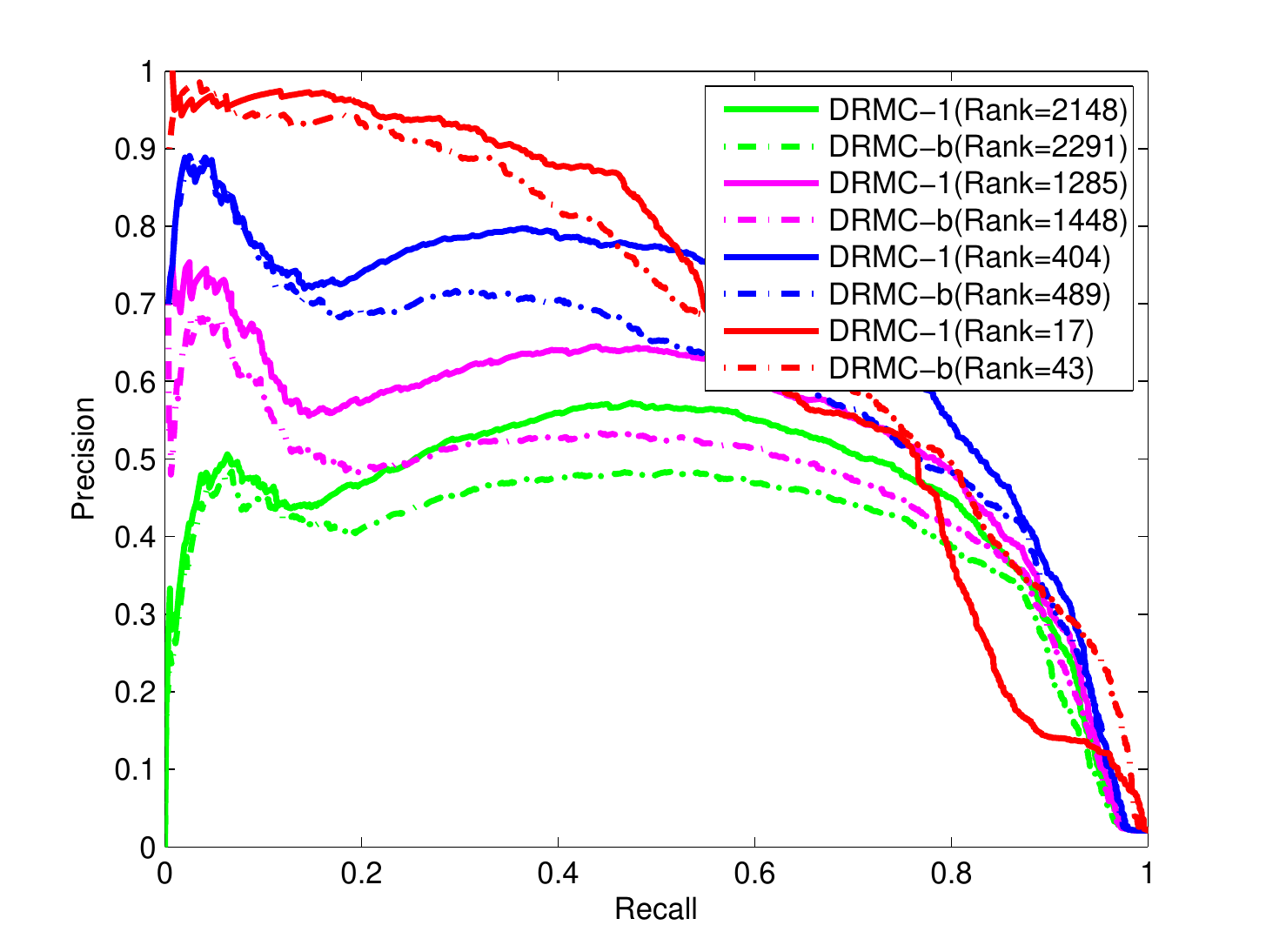}}
  \subfigure{
    \includegraphics[width=0.26\textwidth]{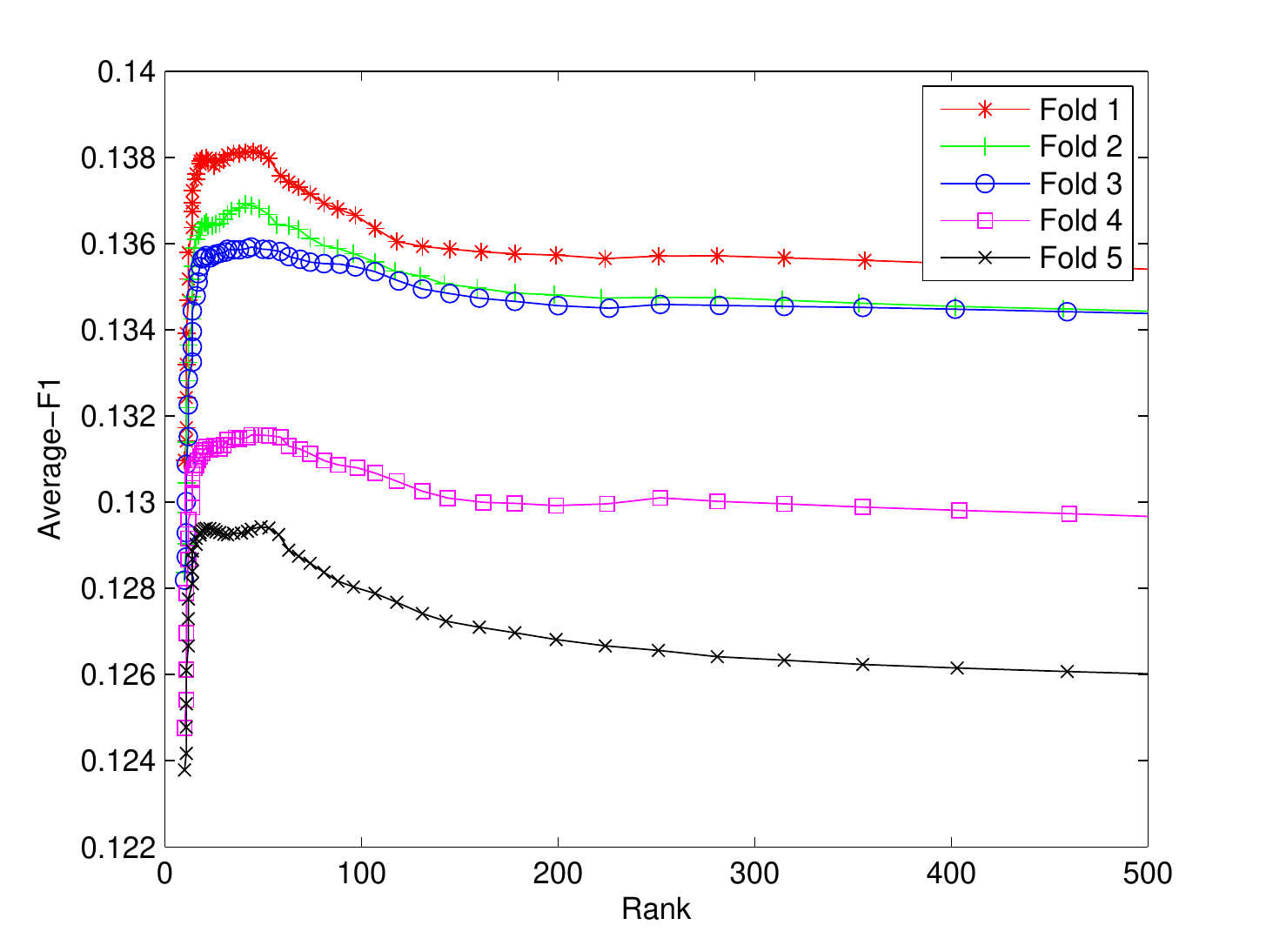}}
    \hspace{-15pt}
  \subfigure{
    \includegraphics[width=0.26\textwidth]{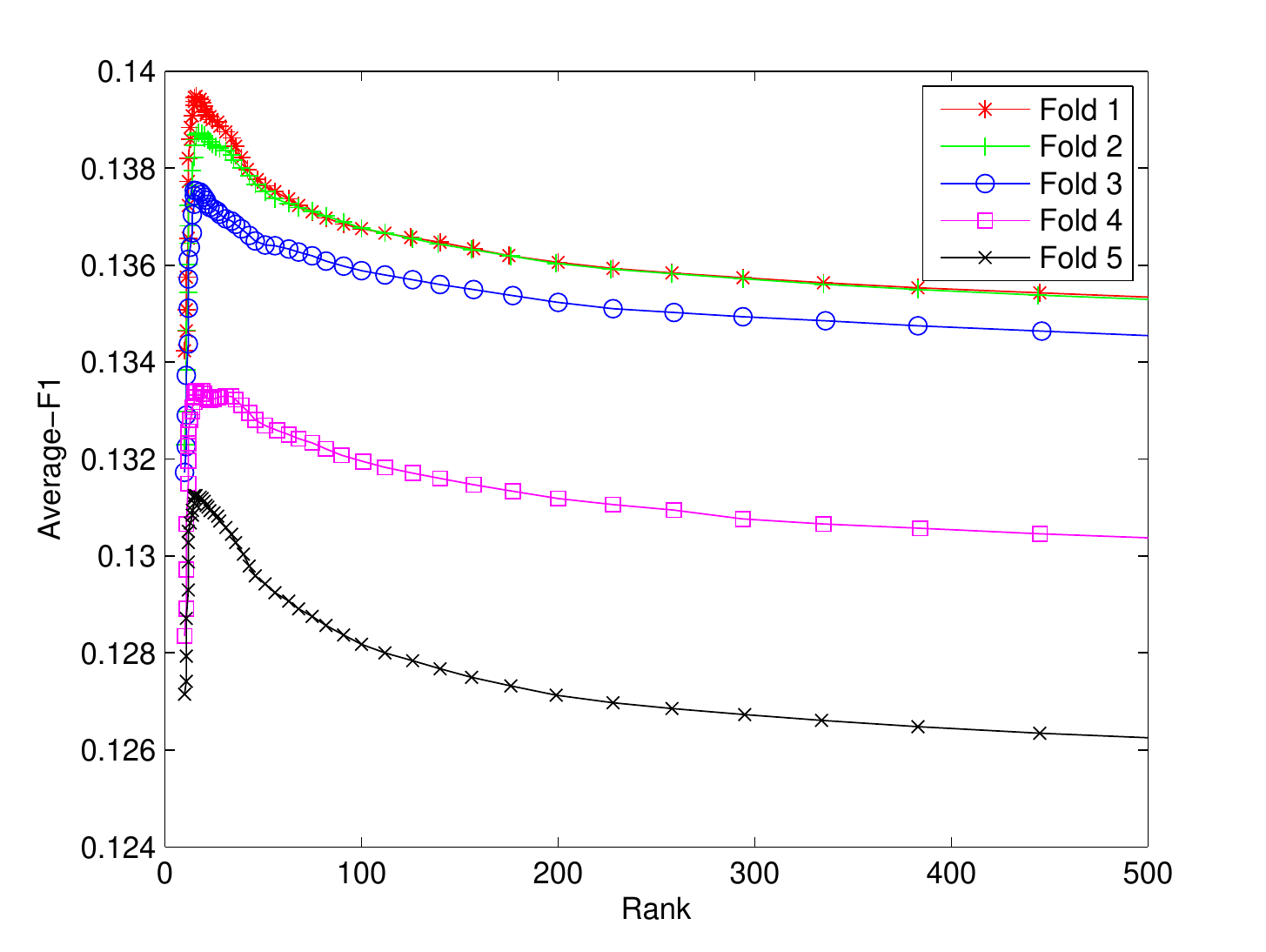}}
    \hspace{-15pt}
  \subfigure{
    \includegraphics[width=0.26\textwidth]{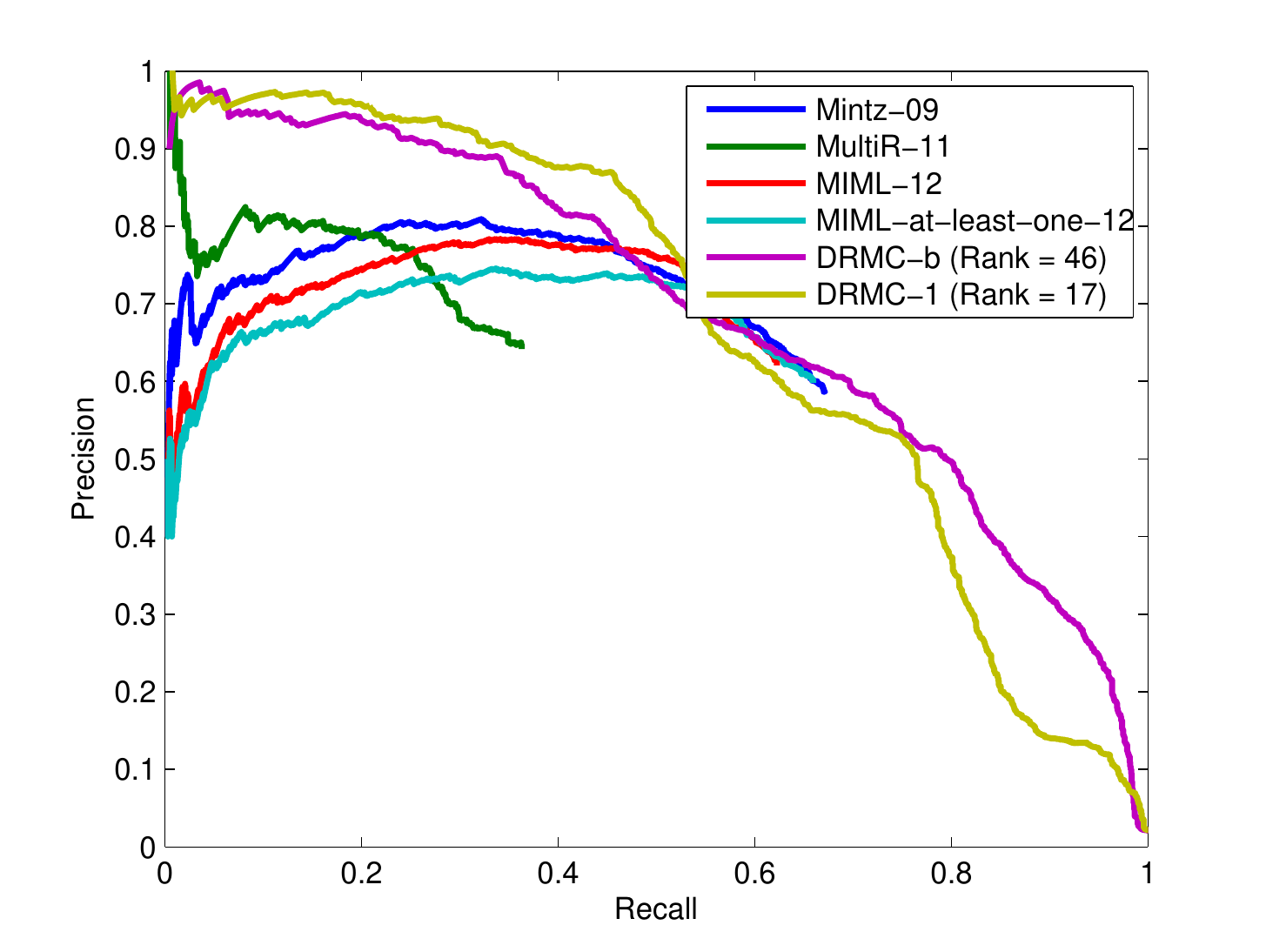}}
    \hspace{-15pt}
  \subfigure{
    \includegraphics[width=0.26\textwidth]{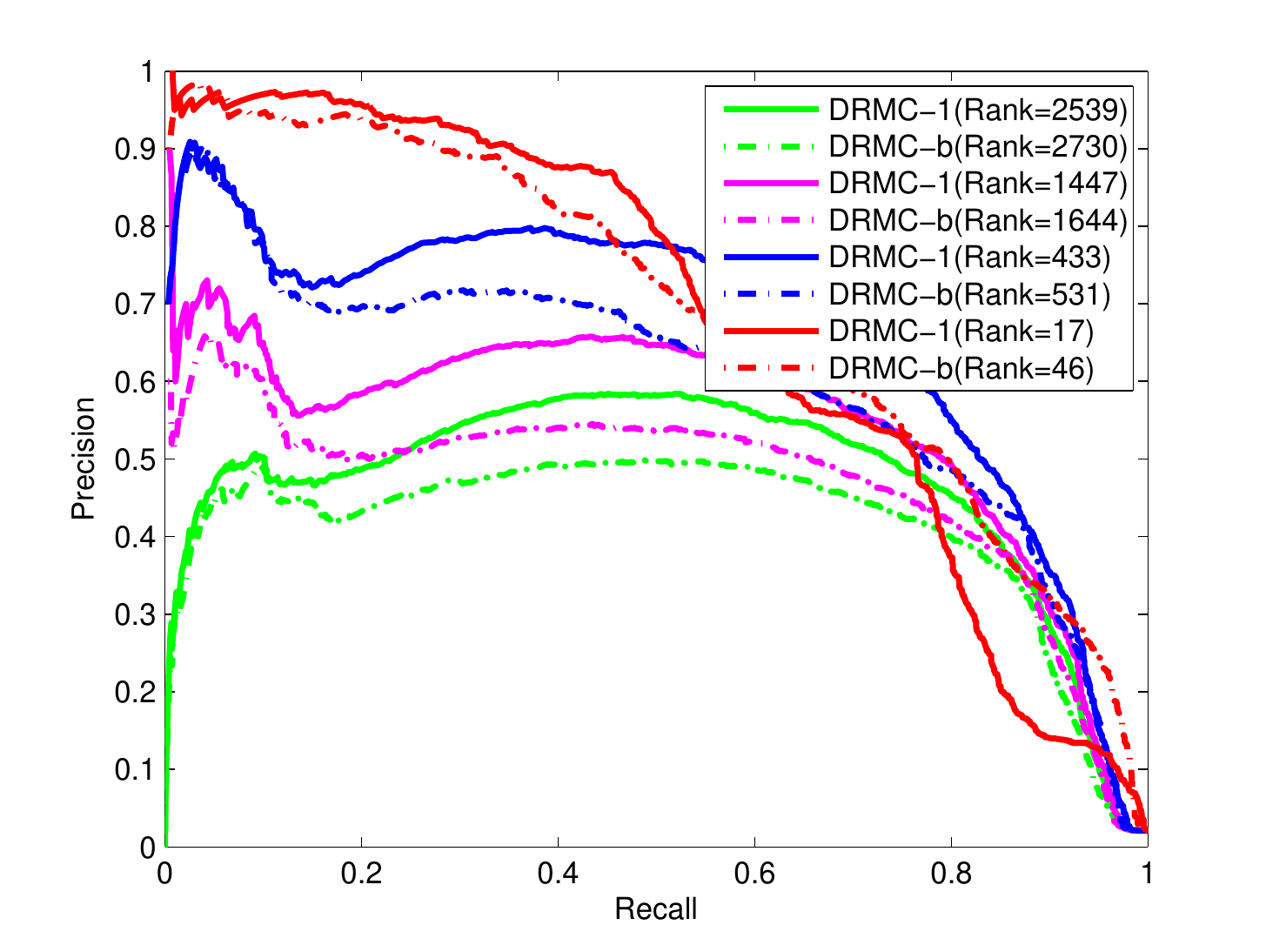}}
  \subfigure{
    \includegraphics[width=0.26\textwidth]{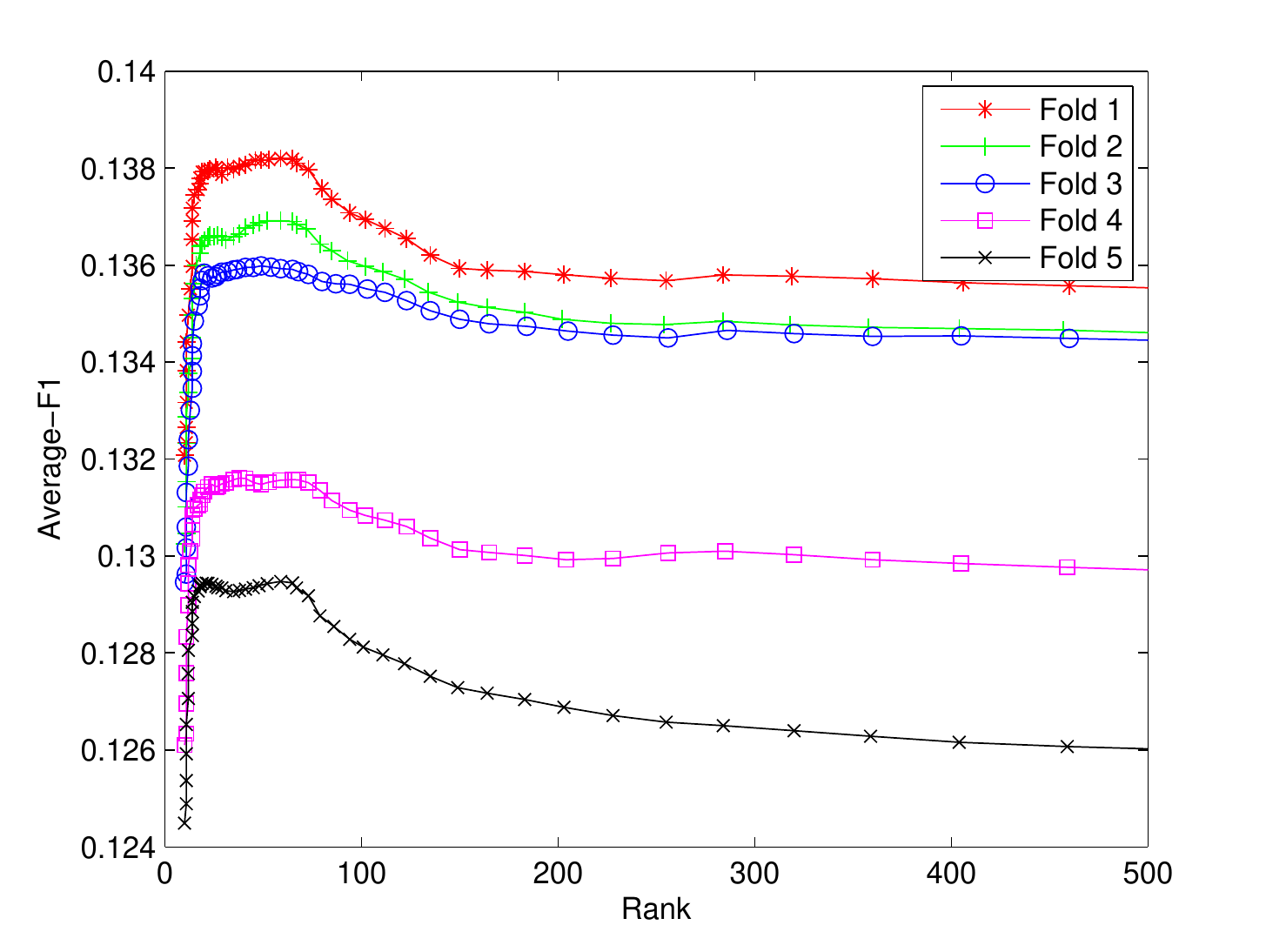}}
    \hspace{-15pt}
  \subfigure{
    \includegraphics[width=0.26\textwidth]{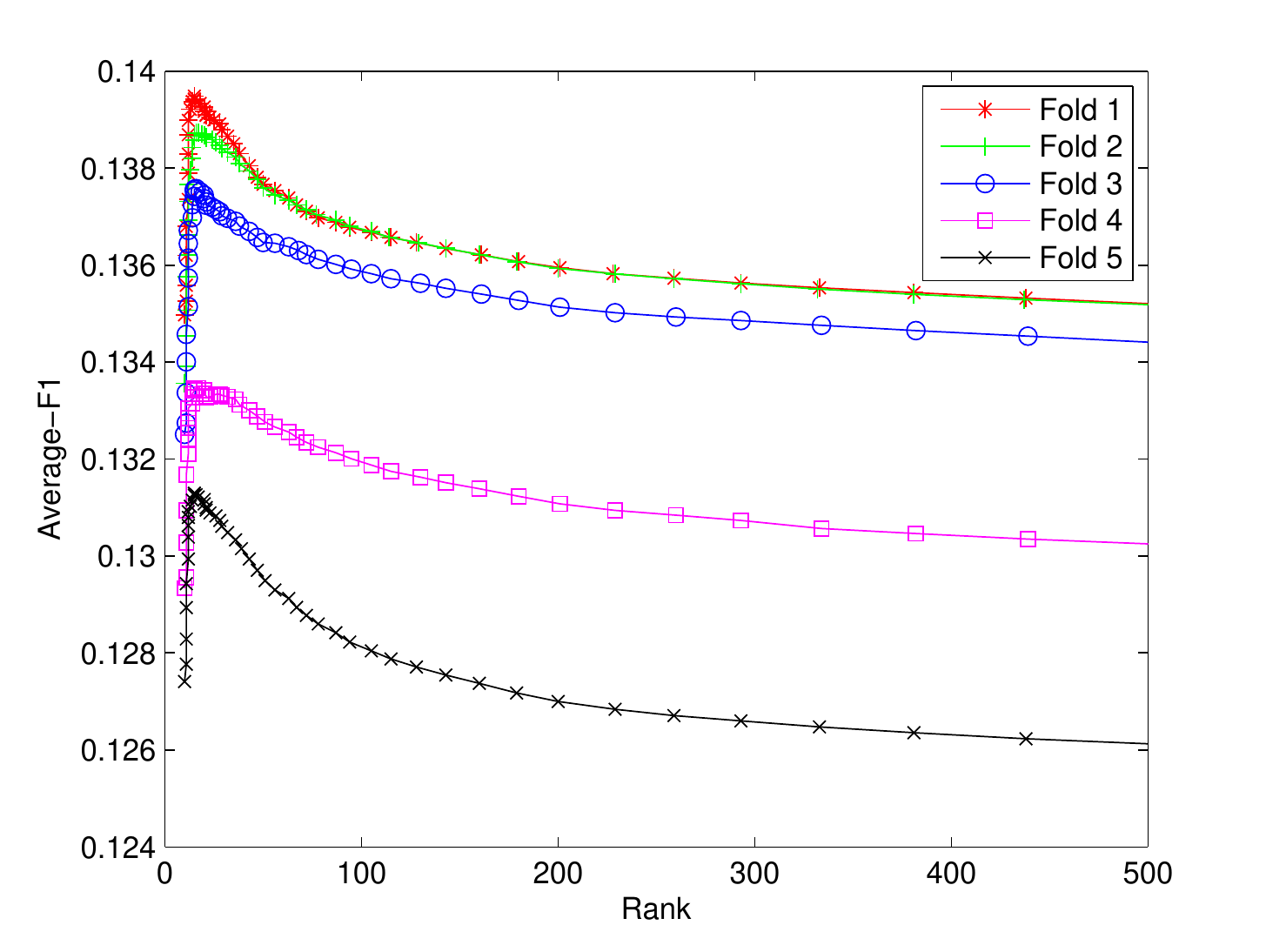}}
    \hspace{-15pt}
  \subfigure{
    \includegraphics[width=0.26\textwidth]{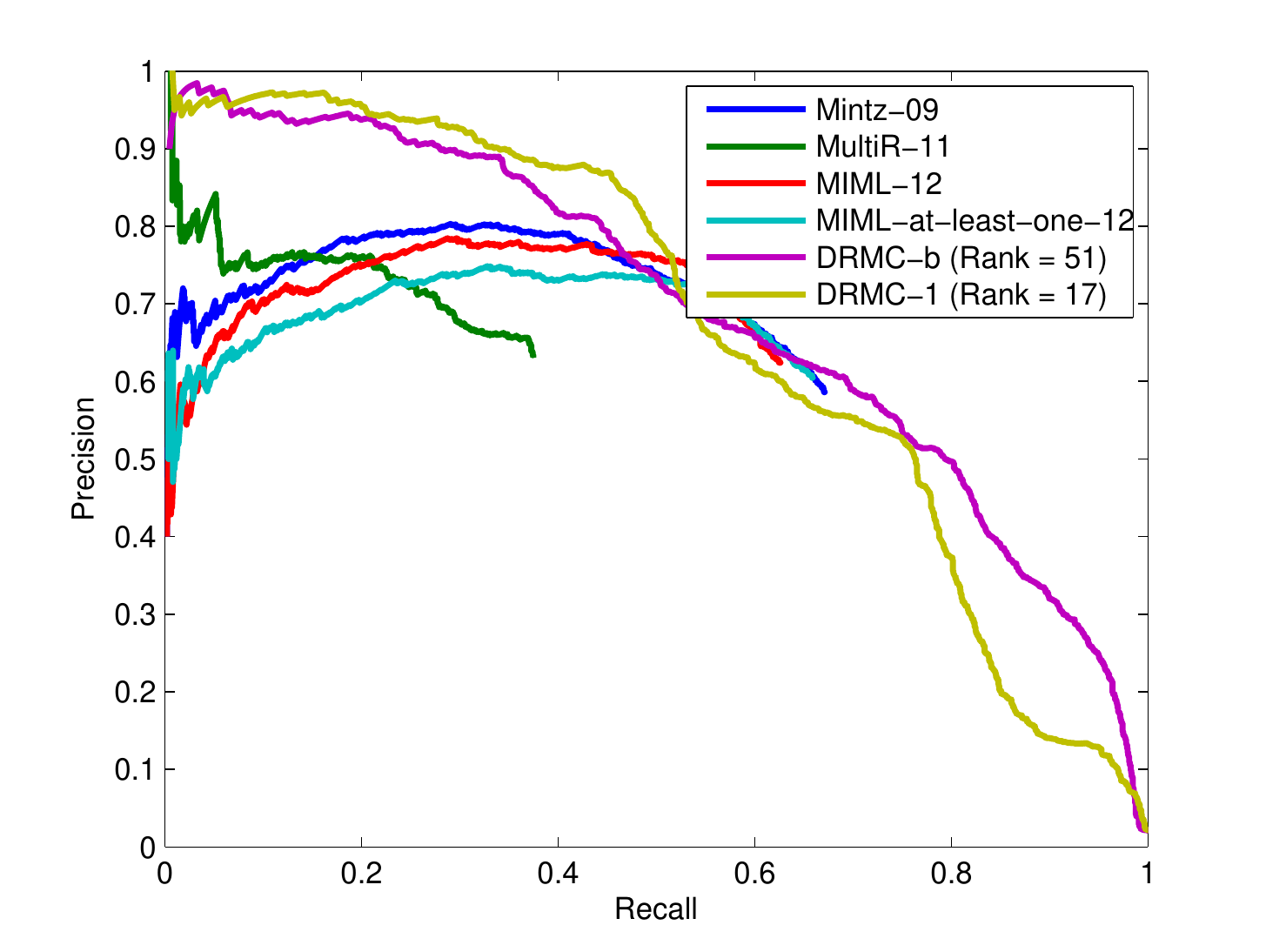}}
    \hspace{-15pt}
  \subfigure{
    \includegraphics[width=0.26\textwidth]{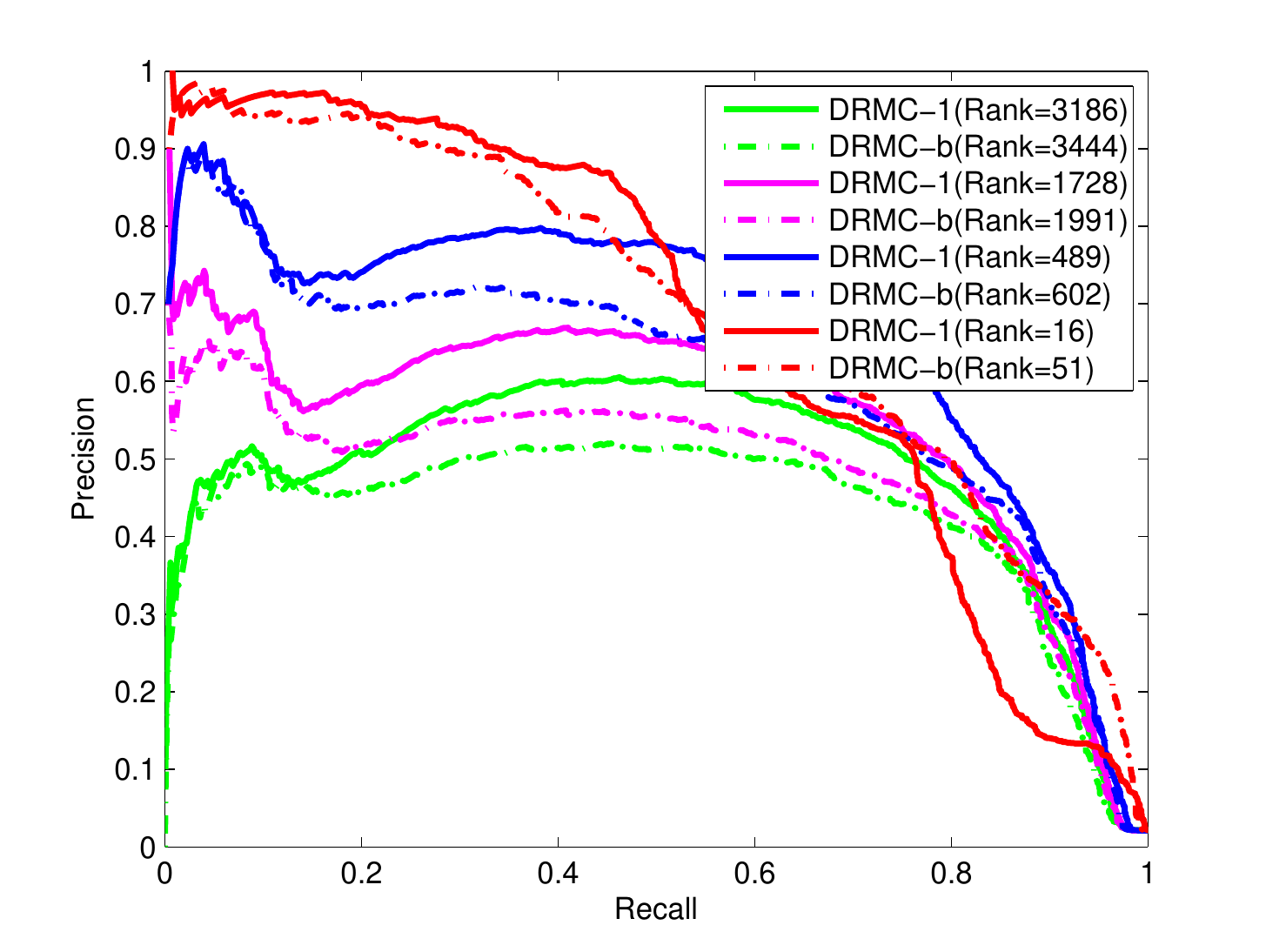}}

  \caption{Feature sparsity discussion on NYT'10 testing set. Each row (from top to bottom, $\theta = 4, 3, 2$) illustrates a suite of experimental results. They are, from left to right, five-fold cross validation for rank estimation on DRMC-b and DRMC-1, method comparison and precision-recall curve with different ranks, respectively.}

\end{figure*}
\begin{table}
\centering
\begin{tabular}{|c|c|c|c|}
  \hline
        Top-N & NFE-13 & DRMC-b & DRMC-1 \\
        \hline
       Top-100 & 62.9\% & {\bf 82.0\%} & 80.0\%\\
       Top-200 & 57.1\% & 77.0\% & {\bf 80.0\%}\\
       Top-500 & 37.2\% & 70.2\% & {\bf 77.0\%}\\
       \hline
       Average & 52.4\% & 76.4\% & {\bf 79.0\%}\\
  \hline
\end{tabular}

\caption{Precision of NFE-13, DRMC-b and DRMC-1 on Top-100, Top-200 and Top-500 predicted relation instances.}
\end{table}
In practical applications, we also concern about the precision on Top-N predicted relation instances. Therefore, We compare the precision of Top-100s, Top-200s and Top-500s for DRMC-1, DRMC-b and the state-of-the-art method NFE-13 \cite{riedel-EtAl:2013:NAACL-HLT}. Table 3 shows that DRMC-b and DRMC-1 achieve 24.0\% and 26.6\% precision increments on average, respectively.

\section{Discussion}
We have mentioned that the basic alignment assumption of distant supervision \cite{mintz2009distant} tends to generate noisy (noisy features and incomplete labels) and sparse (sparse features) data. In this section, we discuss how our approaches tackle these natural flaws.

Due to the noisy features and incomplete labels, the underlying low-rank data matrix with truly effective information tends to be corrupted and the rank of observed data matrix can be extremely high. Figure 5 demonstrates that the ranks of data matrices are approximately 2,000 for the initial optimization of DRMC-b and DRMC-1. However, those high ranks result in poor performance. As the ranks decline before approaching the optimum, the performance gradually improves, implying that our approaches filter the noise in data and keep the principal information for classification via recovering the underlying low-rank data matrix.

Furthermore, we discuss the influence of the feature sparsity for our approaches and the state-of-the-art methods. We relax the feature filtering threshold ($\theta = 4, 3, 2$) in Surdeanu et al.'s \shortcite{Surdeanu2012} open source program to generate more sparse features from NYT'10 dataset. Figure 6 shows that our approaches consistently outperform the baseline and the state-of-the-art methods with diverse feature sparsity degrees.
Table 2 also lists the range of optimal rank for DRMC-b and DRMC-1 with different $\theta$.
We observe that for each approach, the optimal range is relatively stable. In other words, for each approach, the amount of truly effective information about underlying semantic correlation keeps constant for the same dataset, which, to some extent, explains the reason why our approaches are robust to sparse features.

\section{Conclusion and Future Work}
In this paper, we contributed two noise-tolerant optimization models\footnote{The source code can be downloaded from \url{https://github.com/nlpgeek/DRMC/tree/master}}, DRMC-b and DRMC-1, for distantly supervised relation extraction task from a novel perspective. Our models are based on matrix completion with low-rank criterion. Experiments demonstrated that the low-rank representation of the feature-label matrix can exploit the underlying semantic correlated information for relation classification and is effective to overcome the difficulties incurred by sparse and noisy features and incomplete labels, so that we achieved significant improvements on performance.

Our proposed models also leave open questions for distantly supervised relation extraction task. First, they can not process new coming testing items efficiently, as we have to reconstruct the data matrix containing not only the testing items but also all the training items for relation classification, and compute in iterative fashion again. Second, the volume of the datasets we adopt are relatively small. For the future work, we plan to improve our models so that they will be capable of incremental learning on large-scale datasets \cite{chang2011foundations}.
\section*{Acknowledgments}
This work is supported by National Program on Key Basic Research Project (973 Program) under Grant 2013CB329304, National Science Foundation of China (NSFC) under Grant No.61373075 and HTC Laboratory.
\bibliographystyle{acl}

\end{document}